\documentclass[lettersize,journal]{IEEEtran}
\usepackage{amsmath,amsfonts}
\usepackage{algorithmic}
\usepackage{algorithm}
\usepackage{array}
\usepackage[caption=false,font=normalsize,labelfont=sf,textfont=sf]{subfig}
\usepackage{textcomp}
\usepackage{stfloats}
\usepackage{url}
\usepackage{verbatim}
\usepackage{graphicx}
\usepackage{mathtools}
\usepackage{bm}
\usepackage{booktabs}
\usepackage{multirow}
\usepackage{pifont} 
\usepackage{cleveref}
\usepackage{color}
\usepackage[numbers]{natbib}

\newcommand{\size}[2]{{\fontsize{#1}{0}\selectfont#2}}

\definecolor{orange}{cmyk}{0,0.5,0.8,0}

\begin{document}

\title{PnLCalib: Sports Field Registration via\\ Points and Lines Optimization}


\author{Marc Guti\'errez-P\'erez and Antonio Agudo\\
Institut de Rob\`otica i Inform\`atica Industrial, CSIC-UPC, Spain
}

\markboth{{PnLCalib}: Sports Field Registration via Points and Lines Optimization}%
{Shell \MakeLowercase{\textit{et al.}}: A Sample Article Using IEEEtran.cls for IEEE Journals}


\maketitle
\begin{abstract}
Camera calibration in broadcast sports videos presents numerous challenges for accurate sports field registration due to multiple camera angles, varying camera parameters, and frequent occlusions of the field. Traditional search-based methods depend on initial camera pose estimates, which can struggle in non-standard positions and dynamic environments. In response, we propose an optimization-based calibration pipeline that leverages a 3D soccer field model and a predefined set of keypoints to overcome these limitations. Our method also introduces a novel refinement module that improves initial calibration by using detected field lines in a non-linear optimization process. This approach outperforms existing techniques in both multi-view and single-view 3D camera calibration tasks, while maintaining competitive performance in homography estimation. Extensive experimentation on real-world soccer datasets, including SoccerNet-Calibration, WorldCup 2014, and TS-WorldCup, highlights the robustness and accuracy of our method across diverse broadcast scenarios. Our approach offers significant improvements in camera calibration precision and reliability. Our project is available at \url{https://github.com/mguti97/PnLCalib}.
\end{abstract}

\section{Introduction}
\label{sec:intro}
Sports analytics has become an increasingly vital component in modern sports, transforming the way teams, coaches, and fans understand and optimize athletic performance. The proliferation of advanced tracking technologies, such as player and ball tracking systems, has enabled the generation of rich, high-resolution data that provides unprecedented insights into the dynamics of sports competitions. This wealth of tracking data has revolutionized the way sports are analyzed, allowing for more informed decision-making, enhanced player development, and the identification of strategic advantages. Leveraging these data-driven insights has become a key competitive edge, as teams strive to gain a deeper understanding of player movements, team tactics, and in-game patterns. The ability to accurately capture and analyze tracking data has become a crucial aspect of sports analytics, fueling innovations in areas like player performance optimization, injury prevention~\citep{blanchard2019keep}, and the development of advanced coaching strategies~\citep{wang2024tacticai}.

While the proliferation of wearable tracking devices has been instrumental in generating sports performance data, the use of computer vision techniques has emerged as a compelling alternative approach. By leveraging advanced computer vision algorithms, researchers and sports organizations can now extract valuable tracking data directly from video footage, without the need for intrusive wearable sensors~\citep{perez-yuswacv2022}. This camera-based tracking approach offers several advantages, including the ability to capture data from multiple athletes simultaneously, the elimination of potential interference or disconnection issues associated with wearables, and the potential for retroactive analysis of historical game footage. Computer vision-based tracking leverages techniques such as object detection, object tracking, and pose estimation to accurately identify and monitor the movements of players, balls, and other key elements within a sports environment. This data-driven, non-invasive approach to tracking has become increasingly sophisticated, enabling the generation of rich, high-fidelity datasets that can provide deeper insights into athletic performance and team dynamics.

The applications of computer vision in sports extend far beyond just tracking player and ball motions. Innovative computer vision tools have been leveraged to enhance various aspects of the sports experience. One prominent example is the use of semi-automatic offside detection systems, which leverage computer vision algorithms to quickly and accurately determine offside positions during live matches, providing crucial support to referees and improving the fairness and pace of the game. Additionally, computer vision techniques are being employed to generate real-time graphics and overlays for sports broadcasts, seamlessly integrating information such as player statistics, team formations, and tactical visualizations. This enhanced visual experience not only informs and engages the audience but also creates new opportunities for data-driven storytelling and fan engagement. Looking ahead, the continued advancements in computer vision are poised to revolutionize various facets of sports, from automated refereeing and in-depth performance analytics to immersive fan experiences and the integration of augmented reality into the viewing experience.

While the advancements in sports analytics and computer vision have enabled unprecedented insights and experiences, one crucial aspect that underpins these capabilities is the accurate calibration of cameras used to capture sports footage. Camera calibration refers to the process of determining both intrinsic and extrinsic parameters of a camera system, which is essential for transforming 2D image data into meaningful 3D representations of the sports environment. While sports fields, with their well-defined dimensions~\citep{FIFAstadium}, serve as calibration objects, achieving accurate camera
calibration in the broadcast setting poses challenges due to multiple camera views, focal length variability and partial occlusion of the court, hindering the matching process between 2D and 3D correspondences.

Recent developments in camera calibration have moved beyond traditional handcrafted approaches and now encompass a wide spectrum of learning-based techniques tailored to various camera models and deployment scenarios. For standard pinhole cameras, deep learning methods have demonstrated the ability to estimate both intrinsic and extrinsic parameters from image content alone. Early work like DeepFocal~\citep{workman2015deepfocal} focused on focal length regression, while PoseNet~\citep{kendall2015posenet} introduced the use of CNNs for full 6-DoF pose estimation. Subsequent methods have incorporated geometric reasoning through intermediate representations such as vanishing points~\citep{zhai2016detecting} or horizon lines~\citep{workman2016horizon}, enabling more robust and interpretable estimation pipelines. DeepFEPE~\citep{jau2020deep} and NeurVPS~\citep{zhou2019neurvps}, for instance, improve pose estimation by integrating differentiable modules that mimic traditional keypoint detection and geometric constraint solving. These methods benefit from their ability to generalize across scenes without the need for known targets or structured environments.
In parallel, the calibration of distortion models has also advanced significantly. For radial and rolling shutter distortion, deep networks have been used to learn corrective transformations directly from image data, using either parameter regression or dense distortion maps. Techniques like URS-CNN~\citep{rengarajan2017unrolling} and DeepUnrollNet~\citep{liu2020deep} have proven effective for modeling image formation under non-ideal optical conditions. Beyond single-camera settings, cross-view calibration addresses relative pose or homography estimation between images captured from different viewpoints. Approaches such as UDHN~\citep{nguyen2018unsupervised}, BasesHomo~\citep{hold2018perceptual}, and HomoGAN~\citep{hong2022unsupervised} use supervised or unsupervised objectives to learn geometric alignment across diverse perspectives, often without access to ground truth calibration. These are especially useful when image pairs exhibit wide baselines or temporal variation. Cross-sensor calibration further extends the calibration problem to systems involving different modalities—such as RGB cameras and LiDAR. Networks like RegNet~\citep{schneider2017regnet}, CFNet~\citep{lv2021cfnet}, and SemAlign~\citep{liu2021semalign} estimate spatial correspondences by aligning semantic, geometric, or point cloud features across sensors. Moreover, calibration setups that leverage external devices—such as collimator systems~\citep{liang2024camera}—offer a practical alternative in environments where target-based calibration is infeasible due to long working distances or spatial constraints.
Complementing these advances, auto-calibration methods have continued to evolve, offering solutions that do not rely on dedicated calibration targets or full projective reconstructions. Classical approaches ~\citep{triggs1998autocalibration, castorena2016autocalibration} depend on strong assumptions, such as knowledge of multiple fundamental matrices or fixed internal parameters across views. In contrast, newer methods explore more flexible formulations that make use of partially known intrinsics or multi-view image correspondences. Recent work~\citep{cin2024minimal} has further shown that by leveraging minimal geometric constraints.

Traditional sports field registration relied on feature-based methods~\citep{puwein2011robust}, such as detecting and matching local features like SIFT~\citep{lowe2004distinctive} and MSER~\citep{matas2004robust} to estimate pairwise correspondences and compute a homography matrix by using RANSAC~\citep{fischler1981random}. The recent surge in deep learning has led to several data-driven approaches leveraging Convolutional Neural Networks (CNNs) for feature extraction, showing promising results in sports field registration. These methods include field-specific feature prediction~\citep{citraro2020real, nie2021robust, chu2022sports, oo2023residual, homayounfar2017sports, falaleev2024enhancing, Gutierrez-Perez_2024_CVPR} and direct homography matrix regression~\citep{jiang2020optimizing, tarashima2020sflnet}.
Other researchers have investigated camera calibration as a search problem~\citep{sharma2018automated, chen2019sports, sha2020end, zhang2021high, zhang2023four, mavrogiannis2024using}, generating camera pose databases and refining estimates to improve calibration accuracy. Moreover, some approaches~\citep{agudoICPR2020,gupta2011using, ghanem2012robust, citraro2020real, nie2021robust, claasen2023video} leverage temporal calibration consistency between video frames, intending to better align with the nature of sports video broadcasts. Focusing on the soccer domain, despite the potential of estimating camera parameters for reconstructing non-planar points and enabling applications such as automatic camera control, offside detection, or 3D ball tracking, previous studies~\citep{chen2019sports, sha2020end, sharma2018automated, jiang2020optimizing, citraro2020real, zhang2021high, zhang2023four, nie2021robust, shi2022self, chu2022sports, oo2023residual} have predominantly treated the task as homography estimation rather than full calibration~\citep{theiner2023tvcalib, mavrogiannis2024using}.

\begin{figure*}[h]
  \centering
  \includegraphics[scale=0.125]{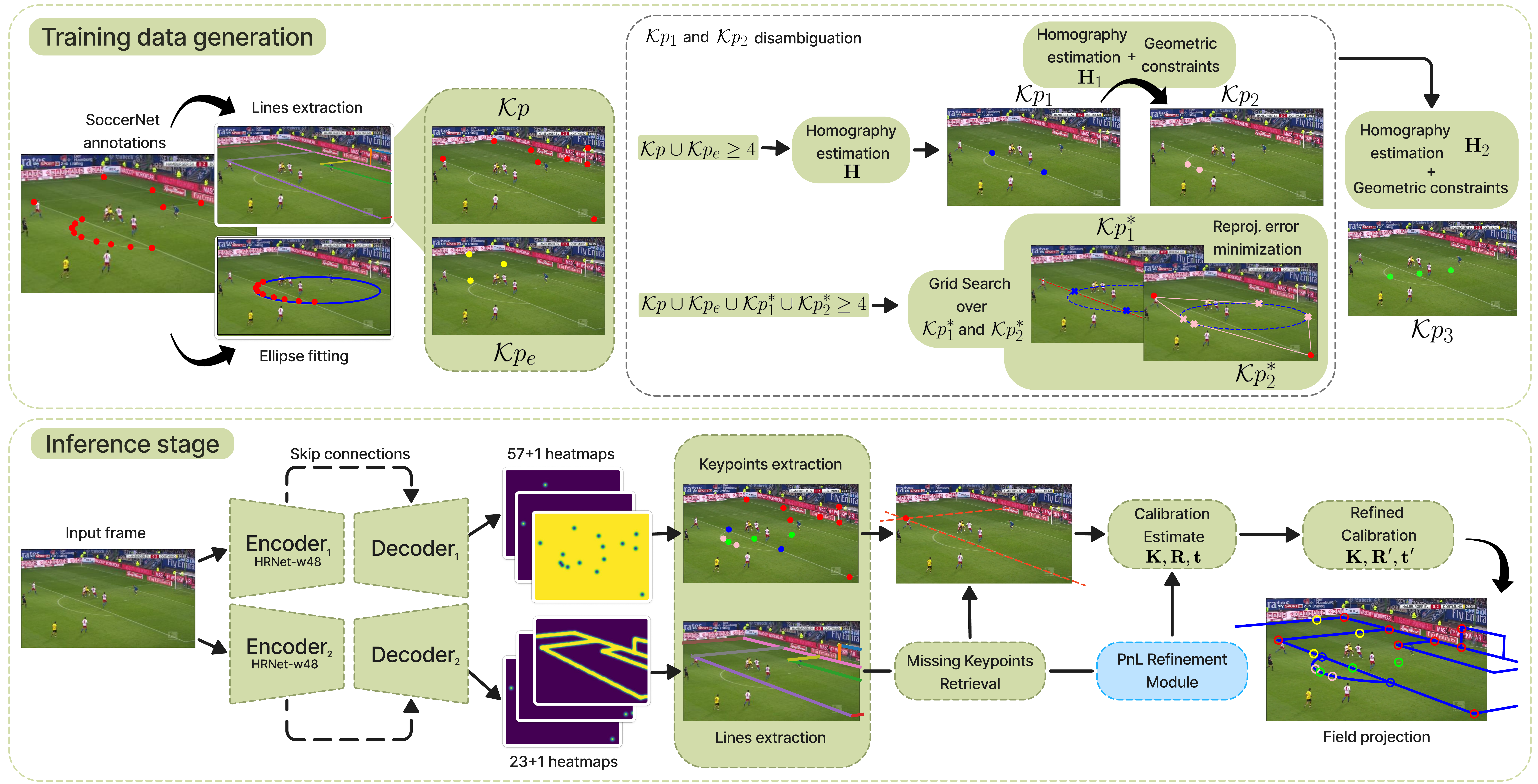} 
  \caption{\textbf{Overview of our proposed framework.} \textbf{Top:} Training data generation pipeline. Beginning with SoccerNet~\citep{cioppa2022scaling} annotations, we utilize field line extraction and ellipse fitting to establish a hierarchical structure for computing each set of keypoints. \textbf{Bottom:} Inference stage pipeline. The encoder-decoder networks produce heatmaps for keypoints and extremities of soccer field lines to extract their positions in the image space. The obtained keypoint set is augmented with intersections of lines generated by the second model to ensure a sufficient number of points. After initial calibration, our PnL refinement module is applied to further refine the calibration estimate by jointly using detected points and lines information.}
  \label{fig:pipeline}
\end{figure*}

Inspired by the limitations of existing approaches, we propose a novel calibration pipeline (see Fig.~\ref{fig:pipeline}) for 3D sports field registration. An early version of this work was presented in~\citep{Gutierrez-Perez_2024_CVPR}, in which we proposed our method to be capable of addressing the challenges posed by the multiple-view broadcast nature. This approach involves defining a hierarchical pipeline to extract a pre-defined keypoint grid from the court's geometric properties and leveraging an encoder-decoder network to estimate keypoint positions. Moreover, the soccer field's lines are defined following the SoccerNet~\citep{cioppa2022scaling} notation and line extremities are also extracted. Particularly, we adopt HRNetv2~\citep{wang2020deep} as the backbone model for the keypoints and line extremities prediction.
The estimated keypoints are used to compute an initial estimate of the projection matrix using RANSAC~\citep{fischler1981random} and Direct Linear Transformation (DLT)~\citep{hartley2003multiple} algorithms. In this paper, we extend our contribution by incorporating a novel refinement module that jointly uses the detected keypoints and lines further to optimize the initial estimate as a non-linear least-squares problem~\citep{triggs2000bundle}. We extensively evaluate our approach on three real-world soccer broadcast datasets, including SoccerNet-Calibration~\citep{cioppa2022scaling}, WorldCup 2014~\citep{homayounfar2017sports}, and TS-WorldCup~\citep{chu2022sports} datasets, and compare it with state-of-the-art methods in both 2D and 3D sports field registration. The experiments demonstrate that our model achieves superior performance on 3D camera calibration while maintaining comparable results on homography estimation with respect to competing approaches.
In summary, this paper makes the following contributions:
\begin{itemize}
\item A novel geometry-based keypoints grid and a robust pipeline for their retrieval.
\item A calibration pipeline capable of integrating non-planar points for 3D camera calibration and extending to multiple views from the broadcast.
\item A refinement module able to optimize the calibration estimate by jointly using the detected keypoints and lines.
\end{itemize}
\section{Related Work}
\label{sec:relatedwork}

Sports field registration is a critical component of most sports applications in computer vision, whose common approaches intend to estimate homography matrices in team sports. Traditionally, homography estimation has relied on identifying corresponding features or keypoints between images and the court field model. These features, typically obtained by exploiting geometric primitives such as lines and/or circles, are subsequently used to estimate the mapping between the images. This is often done using the RANSAC algorithm~\citep{fischler1981random} in conjunction with DLT~\citep{hartley2003multiple} or non-linear optimization techniques~\citep{triggs2000bundle} that minimize a particular loss function. More recent approaches have diverged from this traditional manner. Some directly predict an initial homography matrix, while others seek the optimal matching homography within a reference database containing synthetic images with known homography matrices. Furthermore, the latest approaches have shifted towards directly retrieving camera parameters instead of the homography matrix. This is achieved through various means, including direct prediction, optimization techniques, leveraging databases of image-pose pairs, or decomposing the homography matrix.

\subsection{Search-based Methods} 
A prevalent approach in the field has been generating synthetic data to populate databases with homographies or camera poses paired with corresponding image features. These features are often derived from edge maps or semantically segmented images, which represent key elements of the sports field such as lines, circles, and other distinctive markings. \cite{sharma2018automated} created a synthetic database of edge map-homography pairs. One of the main drawbacks is using normal distribution in the camera pose sampling, which often leads to non-realistic poses. Addressing this issue, \cite{chen2019sports} created a features-pose database by deducing statistics derived from WorldCup 2014 dataset~\citep{homayounfar2017sports}. Those statistics, related to the camera parameters, were used to sample 90,000 poses which will be encoded through a Siamese Network~\citep{1640964} to distinguish different edge maps. During inference, the network extracts the encoding from edge maps and searches database for nearest neighbor candidate pose. By using area-based semantic segmentation of the soccer field instead of edge images, in contrast with previous approaches, \cite{sharma2018automated} also generated an artificial camera pose database based on the possible ranges for pan and tilt angles and focal length parameters. Aiming to solve image-to-image translation problems, \cite{zhang2021high} use improved semantic segmentation by exploiting a conditional generative adversarial network~\citep{isola2017image}. Overall, search-based methods for camera calibration face inherent trade-offs between database size, processing speed, and estimation accuracy. Smaller databases offer faster searches but may compromise initial estimation quality, while larger databases provide better accuracy at the cost of increased computational time. Additionally, these methods often struggle with non-common camera poses frequently encountered in broadcast videos, such as close-ups or oblique angles, which are typically underrepresented in predefined databases. 

\subsection{Optimization-based Methods} 
Alternatively, common approaches also use edge maps, semantically segmented images or information extracted from the field's visual landmarks, like intersection points or lines, to obtain homography matrix or camera parameters from optimization methods. \cite{homayounfar2017sports} classify field lines using a modified VGG network~\citep{simonyan2014very} and extracts their vanishing points reducing the effective number of degrees of freedom (DoF) of the homography from 8 to 4. The field localization problem is formulated as a Markov random field and it requires at least a pair of both vertical
and horizontal lines to estimate the vanishing points. \cite{citraro2020real} make use of a U-Net architecture~\citep{citraro2020real} to jointly detect semantic keypoints corresponding to field's line intersections and player positions. In addition to obtaining the homography matrix from the 2D-3D correspondences, intrinsic and extrinsic camera parameters are subsequently obtained through homography decomposition~\citep{hartley2003multiple}. In order to alleviate the problem of visual landmarks sparsity, \citep{nie2021robust} propose an encoder-decoder network that jointly outputs a grid of keypoints distributed uniformly on the field template and dense template-features in order to further refine the initial keypoint based homography estimate. Similarly, \cite{maglo2022kalicalib} use an encoder-decoder network to predict a perspective-aware keypoint grid. This is, the points nearest to the camera are more spread out than the point
farthest to the camera in order to compensate for the too large distance variations in the image generated by the perspective effect. More recently, \citep{chu2022sports} also use a grid of 91 uniformly distributed
keypoints and formulate its detection problem as an instance segmentation with dynamic filter learning. This is, the convolution filters are generated dynamically, conditioned on the field image and associated keypoint identity. \cite{oo2023residual} use a Residual EfficientNet-Attention UNet architecture to estimate the initial homography matrix using
pre-defined keypoints to register sports fields. The encoder uses the EfficientNetV2~\citep{tan2021efficientnetv2} as the backbone network, and the decoder consists of deconvolution layers with residual blocks, skip connections and attention gates. \cite{theiner2023tvcalib} introduce a differentiable objective function that is able to learn the camera pose and focal length from segment correspondences. Instance segmentation for each visible line or circle segment is achieved with ResNet~\citep{chen2017rethinking} backbone and then, the segment reprojection error induced by the estimated camera parameters is iteratively minimized with a gradient-based method~\citep{acuna2018insights}. Finally, leveraging the temporal consistency from sports videos, \cite{claasen2023video} propose a Bayesian framework. Inspired by recent developments in tracking-by-detection methods, this work proposes a dynamics model that explicitly relates image keypoint positions from one frame to the next through two stages: the first stage consists of a linear Kalman filter, which considers the image keypoints the only part of its state vector, and the second stage incorporates the initial homography estimate to an Extended Kalman Filter~\citep{jazwinski2007stochastic}, with the assumption that the relative homography between frames is small. \cite{falaleev2024enhancing} significantly increase the number of usable points for calibration by exploiting line-line and line-conic intersections, points on the conics, and other geometric features, followed by a DLT optimization. Although the computational cost of the enumerated approaches is lower, optimization-based methods are, overall, not as robust and accurate as search-based methods. Moreover, optimization-based methods are heavily dependent on the landmark detection accuracy.

\subsection{Prediction-based Methods}
Prediction-based approaches use Deep Neural Network-based (DNN) models to calibrate the moving camera in sports. \cite{jiang2020optimizing} train a DNN that directly regresses a homography parameterization given an input frame. Subsequently, a sports field template is warped according to the initial estimate. A concatenation of the input frame and the warped template is fed into a second DNN that estimates the error of the current warping and the estimated homography is accordingly optimized until convergence. \cite{tarashima2020sflnet} propose SFLNet, a CNN single shot regressor that jointly predicts several outputs: a court metric model defined as an 8-dimensional parameter set, which correspond to the homography's DoF, a court semantic segmentation which divides the sports field's spatial layout into divide a frame into court, person, and background regions, and a label adjacency, which comprises adjacency of label pairs in addition to
their presence in an input frame, regularizing the model training via exploiting contextual information. Towards the design of an end-to-end approach, \cite{shi2022self} propose a self-supervised learning method for homography estimation. This work employs a self-supervised data mining method to train the registration network with an image and its edge map by using an iterative estimation
process controlled by a score regression network to measure the registration error. This method is able to obtain competitive results with previous approaches without the need of any labeled data. Recently, \cite{zhang2023four} propose a four-point calibration method. A cGAN is used to generate semantically segmented frames, eliminating foreground objects. Subsequently, a regression network estimates four points from the frames, which will be used to calculate a homography using DLT algorithm, keeping computational cost low. Lastly, extending to the multiple-view nature of the sports broadcast videos, \cite{mavrogiannis2024using} propose a camera calibration method based on synthetic poses. They build a training data generation pipeline with separate flows for the main-camera frames and the rest of camera locations. For the first case, an EfﬁcientNet~\citep{chen2017rethinking} model is trained to take synthetic edge images as input and regress the camera location. For the second case, a perspective transformation is
calculated from four annotated pairs of corresponding points, and the camera location is obtained through homography decomposition. Moreover, a second model is trained on edge maps produced from the poses for each of the possible camera locations to estimate rotation angles and focal length of the camera. Although yielding competitive results on camera calibration and extending it to multiple views, the location-dependent auxiliary model limits the generalization capability and versatility of the method.

\subsection{Homography and Calibration Refinement} Homography refinement is a crucial step in camera calibration, aiming to achieve an even more accurate homography estimation and camera calibration, when necessary. Previous works use one or a combination of the following methods to refine the initial estimate: \cite{puwein2011robust} refine the initial homography by bundle adjustment~\citep{triggs2000bundle}. \cite{chen2019sports} use the Lucas-Kanade algorithm~\citep{baker2004lucas} to refine the initial homography matrix reducing the distance from every pixel in the testing image to its closest edge pixels from a retrieved edge image. \cite{sha2020end} introduce the spatial transformer network (STN) to handle large non-affine transformation. The method stacks the input semantic image and the selected template to feed the STN, which outputs a relative homography that maps the semantic image to the template. \cite{citraro2020real} use the detected players' position on the court's plane to refine the initial homography estimate. After homography decomposition is performed to extract intrinsic and extrinsic camera parameters, these are further refined with Levenberg–Marquardt~\citep{hartley2003multiple} algorithm. A common approach~\citep{zhang2021high, zhang2023four, mavrogiannis2024using} is to use enhanced correlation coefficient method~\citep{evangelidis2008parametric}, which performs image alignment between its inputs and returns the refined homography matrix. Alternatively, \cite{oo2023residual} train an homography refinement network by applying random perturbations to input images. The network outputs the transformation matrix to be applied to the initial estimate. Lastly, other approaches exploit the temporal consistency between subsequent video frames information, such as dense feature maps~\citep{nie2021robust}, homography matrices~\citep{nie2021robust, claasen2023video} or intrinsic and/or extrinsic camera parameters~\citep{citraro2020real}.

\section{Methodology}
\label{sec:methodology}

Sports TV broadcasts consist of video sequences featuring a fraction of the sports field from different uncalibrated moving camera perspectives, defining a multiple-view setting. Our approach focuses on retrieving both extrinsic and intrinsic camera parameters from each individual frame, without any prior information about the camera's position or orientation, except for a partial view of the soccer field. The proposed open-source method comprises five processing components: soccer field modeling and keypoints generation, keypoints and line extremities detection, DLT algorithm and camera parameters retrieval, as well as calibration estimate refinement. Next, these components are introduced.

\begin{figure}[h]
  \centering
  \includegraphics[trim= 125 60 100 68, clip, width=0.49\textwidth]{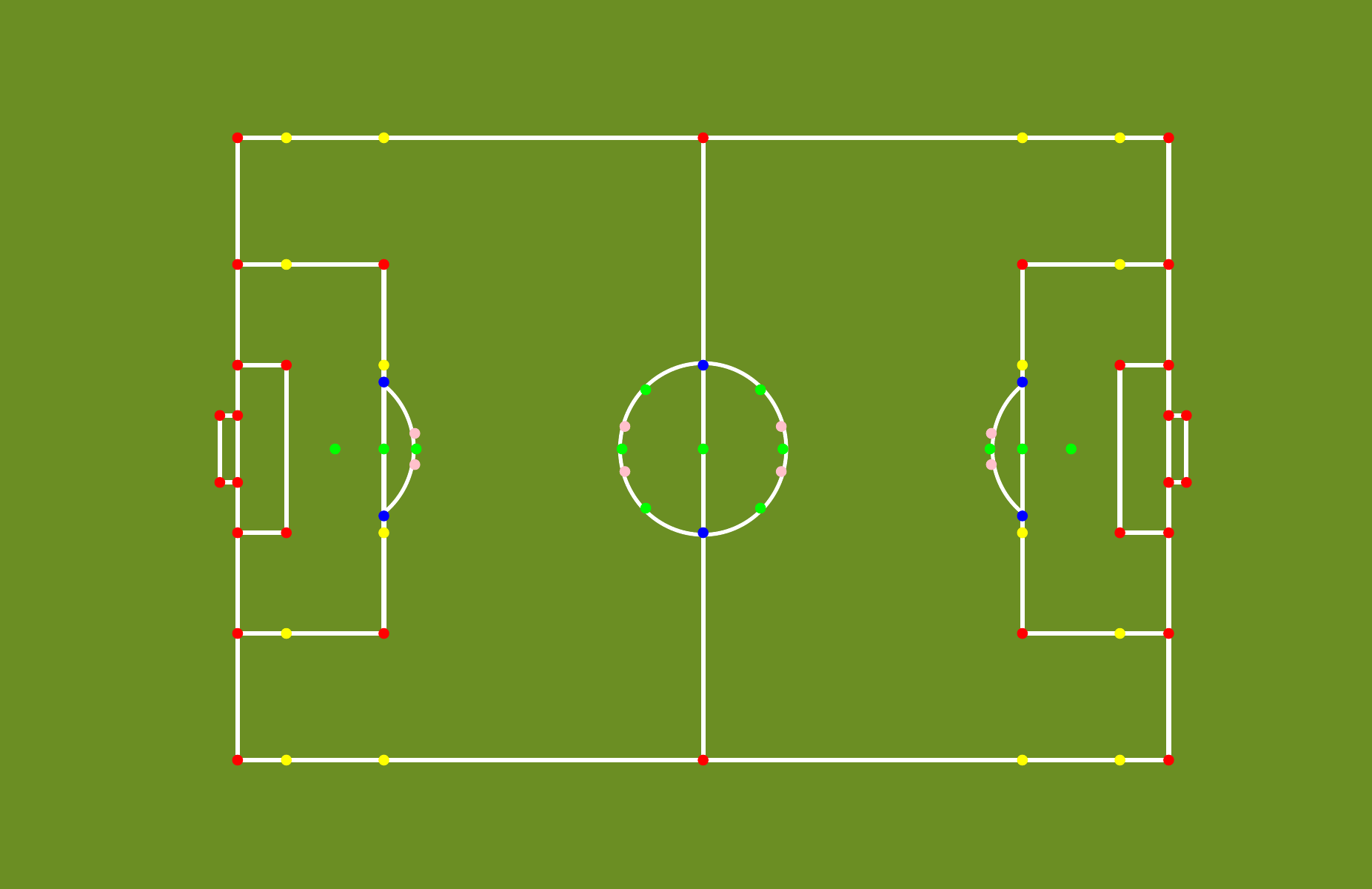} 
  \hfill
  \includegraphics[trim= 55 170 100 135, clip, width=0.49\textwidth]{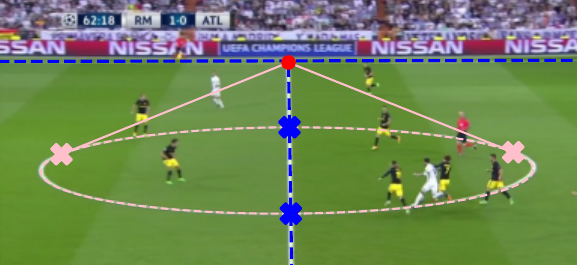}
  \caption{\textbf{Definition of keypoint positions on a soccer field.} \textbf{Top:} Distribution of points on a zenithal view, including all the relevant locations as a result of intersecting lines or curves in the field. $\mathcal{K}p$, $\mathcal{K}e$, $\mathcal{K}p_{1}$, $\mathcal{K}p_{2}$ and $\mathcal{K}p_{3}$ point sets are displayed in red, yellow, blue, pink and green points, respectively. \textbf{Bottom:} Given an external point, both $\mathcal{K}p_{1}$ and $\mathcal{K}p_{2}$ candidates are analytically derived, marked as blue and pink crosses, respectively.}
  \label{fig:fieldmodel}
\end{figure}

\subsection{Modeling the Soccer Field}
A soccer field is composed of lines, circles and semi-circles, representing all field markings, goal posts, and crossbars. Our approach, following keypoint-based methods~\citep{tarashima2020sflnet, citraro2020real, nie2021robust, chu2022sports, maglo2022kalicalib, claasen2023video, oo2023residual, falaleev2024enhancing}, relies on the lines painted on the ground, their intersections, the corners they define and some extra geometric properties, due to its known position on the world coordinate system. We follow the segment definitions of~\cite{cioppa2022scaling} and set them as starting points to hierarchically compute our pre-defined keypoint grid using court geometric properties.

\subsubsection{Keypoint Generation}
The full set of sampled keypoints is organized into subgroups based on the specific geometric features they represent (see Fig.~\ref{fig:fieldmodel}). The hierarchical nature of the keypoint generation pipeline ensures that information from initially identified keypoints is exploited for computing the subsequent ones (some instances in Fig.~\ref{fig:pipeline}-Top). Next, we define the keypoints sets:
\begin{itemize}
    \item \textbf{Line-Line intersections.} Following~\citep{tarashima2020sflnet, cuevas2020automatic, citraro2020real, claasen2023video, oo2023residual, falaleev2024enhancing}, this set of points ($\mathcal{K}p$) includes the intersections of boundary lines, and the penalty or goal area markings. Considering the 23 lines depicted in~\citep{cioppa2022scaling}, including goal posts and crossbars, up to 30 points can be included.
    \item \textbf{Extended Line-Line intersections.} Following~\citep{cuevas2020automatic}, this set ($\mathcal{K}p_e$) addresses the intersections of extended lines that represent non-adjacent segments of the soccer field. To obtain this set, the field lines are extended by exploiting their line equations beyond their original boundaries. It is worth noting that not all non-adjacent line intersections are added; if the lines have to be largely extended, small errors in the line equation would lead to huge deviations in the intersection position. To address this, only intersections located within the image boundaries—or at most up to 50\% of the image size beyond them—are added to the set.
    \item \textbf{Line-Ellipse intersections.} Following~\citep{tarashima2020sflnet, cuevas2020automatic, citraro2020real, claasen2023video, oo2023residual, falaleev2024enhancing}, this set ($\mathcal{K}p_{1}$) considers the intersections between the field lines and the circles or semi-circles present on the court. Given the distortions introduced by the camera perspective, conics on the field are considered ellipses for its equation computation. The parameters of these ellipses are fitted using the least squares method~\citep{halir1998numerically}. Line-ellipse intersection points are analytically derived using ellipse and line formulas (see Fig.~\ref{fig:fieldmodel}-Bottom for a visual example).
    \item \textbf{Ellipse tangent points.} Following~\citep{gupta2011using, falaleev2024enhancing}, the augmentation of available points is achieved through the utilization of tangent points on tangent lines, extending from a specified external point to the previously defined ellipses. These tangent points (denoted by $\mathcal{K}p_{2}$) were analytically determined by employing an ellipse equation and incorporating the known coordinates of an external point (see Fig.~\ref{fig:fieldmodel}-Bottom for a visual example).
    \item \textbf{Additional points.} Following~\citep{chu2022sports, claasen2023video, maglo2022kalicalib, nie2021robust, oo2023residual, falaleev2024enhancing}, once the previous keypoint sets and the corresponding homography are inferred, for the sake of grid completeness, an additional set ($\mathcal{K}p_{3}$) of nine points is integrated along the central pitch axis, encompassing the pitch center and penalty points. Additionally, four points are strategically placed to designate quarter turns along the central circle. Furthermore, the homography facilitates the inclusion of other points that are initially missing, addressing situations such as unannotated or wrongly annotated lines.

\end{itemize}

\subsubsection{Keypoint Disambiguation}
Due to the multi-view nature of the SoccerNet dataset~\citep{deliege2021soccernet} and, for instance, considering one of the soccer field's semi-circles, ambiguity appears in its respective $\mathcal{K}p_{1}$ and $\mathcal{K}p_{2}$ keypoints candidates, as shown in Fig.~\ref{fig:fieldmodel}-Bottom. This is, while the expected locations of keypoint pairs in the image are known, the challenge lies in uniquely identifying and matching each individual keypoint. To solve that, we define two different strategies to handle that disambiguation depending on the total number of keypoints generated in the previous sets: when there are sufficient points in the $\mathcal{K}p\cup \mathcal{K}e$ set to infer a homography ($\mathbf{H}$), i.e., four points, $\mathcal{K}p_{1}$ is computed first by choosing the candidates combination that minimizes the reprojection error. Then, we include $\mathcal{K}p_{1}$ to newly infer a homography estimation ($\mathbf{H_1}$) and repeat the same strategy on the $\mathcal{K}p_{2}$ set (obtaining $\mathbf{H_2}$), as it is shown in Fig.~\ref{fig:pipeline}. Otherwise, we perform a grid-search involving both $\mathcal{K}p_{1}$ and $\mathcal{K}p_{2}$ candidates when $\mathcal{K}p\cup \mathcal{K}p_{e}\cup \mathcal{K}p_{1}^*\cup \mathcal{K}p_{2}^*\geq 4$, where $*$ denotes a possible candidate combination. The grid-search iterates over all keypoints candidates in a set-wise manner to avoid unfeasible combinations and keeps the one with minimum reprojection error. In this case, $\mathbf{H_1}$ estimation is bypassed, and $\mathbf{H_2}$ is computed only after resolving both $\mathcal{K}p_{1}$ and $\mathcal{K}p_{2}$. It is worth pointing out that no more keypoint sets beyond $\mathcal{K}p\cup\mathcal{K}p_e$ are computed if neither of the above stated conditions are met. Moreover, once we compute all the pre-defined keypoint sets, two additional geometrical constraints are applied in case that homography estimation or ellipse fitting is not accurate enough. Initially, we manually establish a reprojection error threshold to validate keypoints. Subsequently, through iteration over combinations of keypoints, we construct vectors and ensure that the cross-products maintain consistent signs in both world and image coordinates. This final step is essential in cases where two distinct combinations yield valid top- and bottom-view perspectives of the field while exhibiting identical reprojection errors. Utilizing cross-products enables us to differentiate and retain the keypoint combination corresponding to the field's top-view. The full keypoint generation process is depicted in Fig.~\ref{fig:pipeline}-Top.

\subsubsection{Left-Right Disambiguation}
In sequences where the camera angle aligns with the longitudinal axis of the court, an ambiguity arises regarding the distinction between the right and left halves of the field. Hence, a critical step to ensure consistency and robustness across keypoints and lines detection processes involves differentiating between the two sides. Taking into account the camera calibration evaluation protocol in~\citep{cioppa2022scaling}, which considers both of the ambiguous options and keeps the highest score, this is accomplished by implementing a remap to the ground-truth (GT) values, ensuring that the goal area closest to the camera consistently represents the left side. The process of checking whether or not the mapping should be applied is defined in a heuristic fashion. We compute angles of horizontal and vertical soccer field lines, respectively; and then set a threshold taking into account angle distribution and visual inspection. This approach facilitates an effective model training process by deferring the disambiguation task until after the inference stage as an extra step if needed.

\subsection{Keypoints and Lines Detection}
Our approach is built upon \cite{falaleev2024enhancing} solution, which makes use of two encoder-decoder convolutional neural networks to estimate the position of the pre-defined keypoints and the soccer field lines depicted in~\citep{cioppa2022scaling} excluding conics. While in~\citep{falaleev2024enhancing} the line model is given an auxiliary role to enhance keypoint completeness, in our approach is used as the key component of our refinement module in the last calibration stage. During inference, the former produces heatmaps for each pre-defined keypoint with a single $2$-pixel Gaussian peak sigma positioned in the keypoint location, accompanied by an additional background channel. This additional channel reflects the inverse of the maximum value among the other target feature maps, ensuring that the resultant target tensor behaves as a probability distribution function at every spatial point. Meanwhile, the latter network produces heatmaps for each visible soccer field line within the frame, assigning two Gaussian peaks at the line extremities' locations. Additionally, we introduce an extra channel, known as the boundary channel, to our heatmap following the approach outlined in~\citep{wang2019adaptive}. This augmentation aims to enhance the efficient capture of global information regarding the soccer field and improve extremities detection, particularly near image borders. We effectively extract the position of keypoints and line extremities from the generated heatmaps by employing a max pooling operation, which calculates the maximum value for patches of a feature map, drawing inspiration from the methodology proposed in~\citep{zhou2019objects}. This process is summarized on Fig.~\ref{fig:pipeline}-Bottom.

\subsubsection{Architecture}
Following~\citep{falaleev2024enhancing}, the keypoint and line extremities detection utilized a modified HRNetV2-w48~\citep{wang2020deep} network as the encoder's backbone network. HRNetv2~\citep{wang2020deep} is a new family of convolutional networks that maintains high-resolution representations through the whole process resulting in semantically richer and spatially more precise representations. To improve the spatial resolution of the predicted heatmaps, we incorporated $2\times$ upsampling and concatenated skip-connection features from the corresponding resolution of the convolution stem to fuse the features at different scales. The resulting feature maps are produced at half the spatial resolution of the input image, striking a balance between computational efficiency and spatial detail retention. Additionally, a more lightweight version of the backbone is achieved by reducing the dense representation layers sizes in the network's last stages. The final predictions exhibit half the resolution of the original image, with softmax and sigmoid employed as the final activation functions for keypoints and lines, respectively. 

\subsubsection{Keypoints Mask}
When homography was unavailable due to a limited number of points, the heatmaps associated with points belonging to $\mathcal{K}p_{1}$, $\mathcal{K}p_{2}$, and $\mathcal{K}p_{3}$—which would have been derived from the homography itself—were masked out from the loss function as long as the line to which they belong is included in the GT annotation. Additionally, when the external point required to compute ellipse tangent points in $\mathcal{K}p_{2}$ is not present in $\mathcal{K}p$, the pair of tangent points candidates is also masked out.

\subsection{Camera Projection Model}
We employ a standard full-perspective camera model as:
\begin{equation}
\label{eq:projection}
    \mathbf{P} = \mathbf{KR}[\mathbf{I}\,\vert\,\mathbf{-t}] \in \mathbb{R} ^{3\times4},
\end{equation}
where $\mathbf{R}\in\mathbb{R}^{3\times3}$ and $\mathbf{t} \in \mathbb{R}^3$ denote the extrinsic parameters (rotation and translation, respectively) to map from scene coordinates to camera ones; and $\mathbf{K}\in\mathbb{R}^{3\times3}$, which includes the intrinsic parameters to transform from camera coordinates to image ones. The latter has the form:
\begin{equation}
\label{eq:calibration}
    \mathbf{K} = \begin{bmatrix}
                    \alpha_x & s & x_0\\
                     & \alpha_y & y_0\\
                     & & 1
                    \end{bmatrix} ,
\end{equation}
where $\{\alpha_x,\alpha_y\}$ denote the focal length, $\{x_0, y_0\}$ the principal point, and $s$ the skew value. Following~\citep{hartley2003multiple}, we assume zero skew and a known pixel aspect ratio. Additionally, for simplicity, we assume the principal point ($x_0, y_0$) coincides with the center of the image, i.e., we neglect astigmatism or distortions.

\subsubsection{Camera Parameters Estimation}
Extrinsic and intrinsic parameters in Eq.~\eqref{eq:projection} are inferred by leveraging the coordinates of 3D object points and their corresponding 2D projections using the soccer field model as a calibration rig, following~\citep{zhang2000flexible}, which consists of a closed-form solution followed by a non-linear refinement based on the maximum likelihood criterion. To calibrate the 3D soccer field model rig, we consider two additional vertical planes containing the goal polygons, including non-planar points such as keypoints belonging to the goal posts and crossbars. Additionally, we extend this strategy to enhance completeness by providing estimations when insufficient points are on the ground plane, calibrating over the vertical planes, and subsequently transforming camera parameters to the ground-plane coordinate system. When a sufficient number of points, i.e., 6 keypoints, is available, an initial estimate of the camera's intrinsic matrix $\mathbf{K}$ is inferred. Camera calibration is then performed using this estimation, resulting in a more robust and stable calibration. Otherwise, calibration is done by jointly optimizing all three unknowns: $\mathbf{K}$, $\mathbf{R}$, and $\mathbf{t}$.
To account for keypoint misdetections and other complexities in camera parameter retrieval, such as frames with only one non-planar keypoint visible, the calibration process was repeated on several subsets of keypoints. Similar to~\citep{falaleev2024enhancing}, these subsets were selected based on various heuristics: \textit{full-keypoints}, including all keypoints sets $\mathcal{K}p$, $\mathcal{K}p_e$, $\mathcal{K}p_1$, $\mathcal{K}p_2$ and $\mathcal{K}p_3$; \textit{main-keypoints}, comprising only line-line intersections from the original SoccerNet annotations~\citep{cioppa2022scaling}; and \textit{ground-plane-keypoints}, which excludes non-planar keypoints. Furthermore, we applied a grid of RANSAC~\citep{fischler1981random} reprojection error thresholds to each subset. The final camera calibration values were determined through a heuristic voting process, prioritizing the method yielding a lower reprojection error, with emphasis on the \textit{full-keypoints} subset. 

\subsubsection{Homography Estimation}
Assuming the world coordinate system such that $z = 0$ corresponds to the ground plane, the ground-to-image homography $\mathbf{H}$ can be obtained from the first, second, and fourth columns of the camera projection matrix $\mathbf{P}$ as: 
\begin{equation}
      \mathbf{H} \cong \mathbf{KR}\begin{bmatrix}
    1 & 0 & -t_x\\
    0 & 1 & -t_y\\
    0 & 0 & -t_z
   \end{bmatrix} \in \mathbb{R}^{3\times3} ,
\end{equation}
where $\mathbf{t}=[t_x, t_y, t_z]^{\top}$. 
Nevertheless, inaccurate estimations for keypoints associated with the non-planar rig, such as those belonging to the goalposts and crossbars, may result in a flawed homography estimation. To address this issue, we employ classical homography estimation with DLT~\citep{hartley2003multiple} and RANSAC~\citep{fischler1981random} on the \textit{ground-plane-keypoints} subset. We define a maximum allowable reprojection error to consider a point pair as an inlier and subsequently refine the initial homography estimation matrix using the Levenberg-Marquardt method~\citep{hartley2003multiple} on the 2D-3D point correspondences. Similarly to the approach used for camera parameter estimation, we applied a grid search over RANSAC reprojection error and employed the heuristic voting method, but in this instance, it is restricted to the \textit{ground-plane-keypoints} subset.

\begin{figure}[t!]
  \centering
  \includegraphics[width=0.49\textwidth]{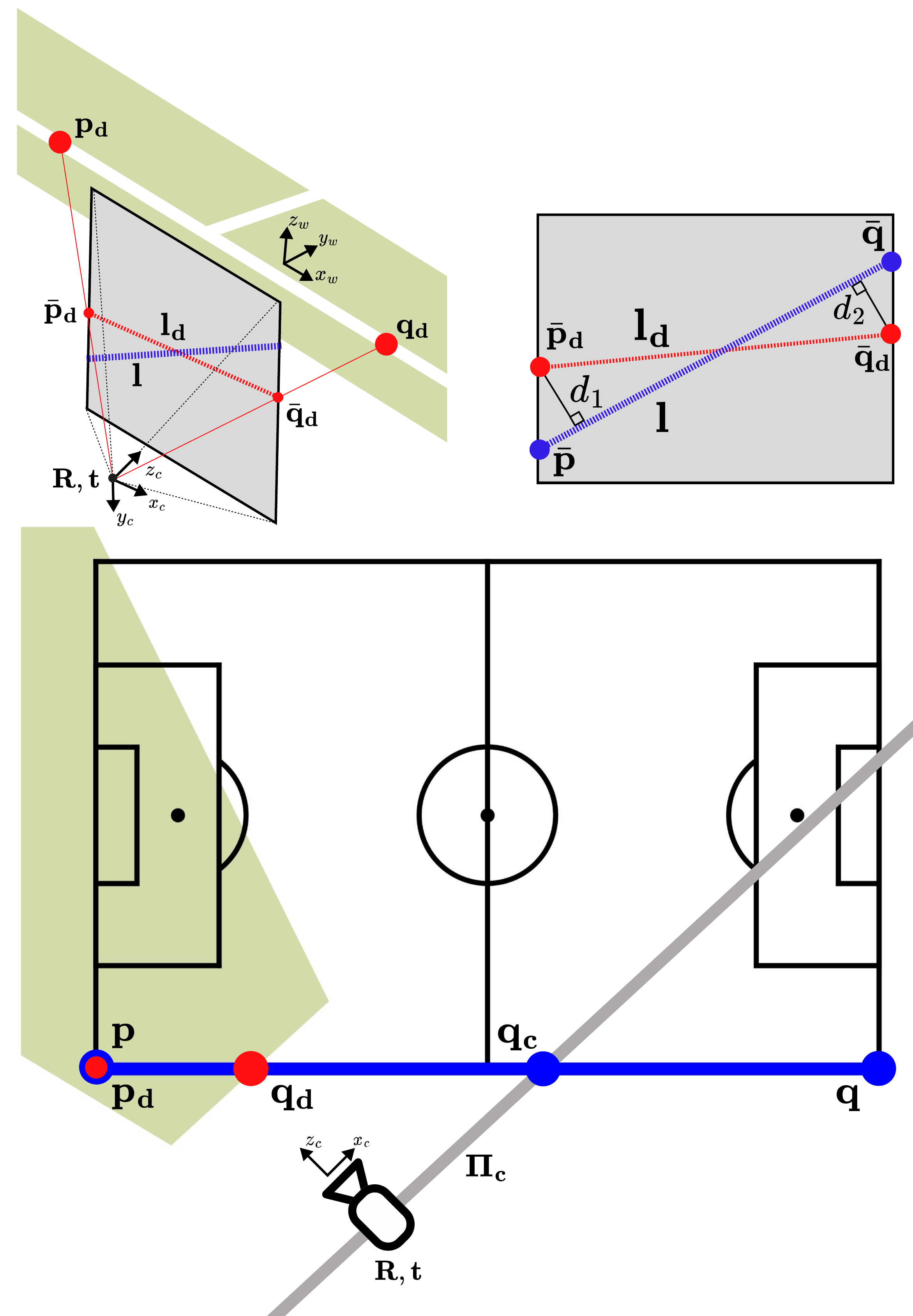} 
  \caption{\textbf{PnL refinement module. Top-Left:} Let $\{\mathbf{p_d},\mathbf{q_d}\}$ be the visible extremities of a detected 3D soccer field line and $\{\mathbf{\bar{p}_d},\mathbf{\bar{q}_d}\}$ their projected 2D points in the image plane, which define the projected line $\mathbf{l_d}$ as well as the detected line projection given an initial calibration estimate. \textbf{Top-Right:}  $d_1$ and $d_2$ represent the line-based reprojection error between $\mathbf{l_d}$ (in red) and $\mathbf{l}$ (in blue). \textbf{Bottom:} Soccer field's top-view. Estimated camera pose $\{\mathbf{R}, \mathbf{t}\}$ defines the camera plane $\mathbf{\Pi_c}$. Let $\{\mathbf{p},\mathbf{q}\}$ be the real extremities of a known 3D line given a field model, and $\mathbf{q_c}$ the corrected 3D point through the line-camera plane intersection. Note that $\mathbf{p}\equiv\mathbf{p_d}$ since the real line extremity is visible in the camera frame. The green area corresponds to the camera viewing cone.}
  \label{fig:linerefine}
\end{figure}

\subsection{Point and Line Refinement Module} 
To address the sparsity of keypoints on the grid, particularly in regions distant from the penalty boxes, where the majority of keypoints are concentrated, and to mitigate issues arising from keypoint misdetections overlapping with detected lines, we have developed a refinement module which makes use of the detected lines information. This module, termed the Point and Line (PnL) refinement module, enhances both homography and calibration estimates by jointly leveraging the information from detected keypoints and lines. The PnL module integrates line information to complement the keypoint data, providing a more robust and comprehensive basis for accurate field registration and camera calibration, especially in areas where keypoint information alone may be insufficient or unreliable. Setting the internal camera calibration matrix $\mathbf{K}$ as a fixed term, PnL refinement module will optimize the calibration estimate through the camera pose $\bm{\Theta} = \{\mathbf{R, t}\}$ space, which includes the rotation and translation parameters, respectively.

Inspired by previous approaches~\citep{pumarola2017pl, vakhitov2016accurate, gomez2019pl}, we next describe the line parameterization, the error function, and its integration within the PnL refinement module. Let $\mathbf{p}, \mathbf{q} \in \mathbb{R}^{3}$ represent the real extremities of a known 3D soccer field line based on a field model, and let $\mathbf{p_d}, \mathbf{q_d} \in \mathbb{R}^{3}$ denote the visible extremities of the same 3D soccer field line detected in the image frame, with image coordinates $\bar{\mathbf{p}}_d, \bar{\mathbf{q}}_d \in \mathbb{R}^2$ defining the detected soccer field line $\mathbf{l}_d = \overrightarrow{\bar{\mathbf{p}}_d\bar{\mathbf{q}}_d}$. Note that if the real line extremities are visible in the camera frame, they are equivalent to the detected extremities, i.e., $\mathbf{p} \equiv \mathbf{p_d}$. Otherwise, the 3D positions of $\mathbf{p_d}$ and $\mathbf{q_d}$ are unknown.

By projecting $\mathbf{p}$ and $\mathbf{q}$ into the image plane, given an initial calibration estimate $\bm{\Theta}$ (or a homography matrix $\mathbf{H}$ in the case of planar estimation), we define the projected soccer field line $\mathbf{l}$ and compute both $\bar{\mathbf{p}}$ and $\bar{\mathbf{q}}$ as its intersections with the image boundaries (see Fig.~\ref{fig:linerefine}-Top). Hence, analogously to the detected lines, $\mathbf{l} = \overrightarrow{\bar{\mathbf{p}}\bar{\mathbf{q}}}$.

However, it is worth noting that 3D line extremities $\mathbf{p}, \mathbf{q}$ lying behind the field's camera plane $\mathbf{\Pi_c}$—i.e., points whose image projections have $z_c < 0$—cannot be projected. To address this issue, we construct the field's camera plane using the estimated camera position $\mathbf{t}$ and a normal vector $\mathbf{n_t}$ to the camera plane (see Fig.~\ref{fig:linerefine}-Bottom). The latter is built based on the camera position and the projection of the principal point onto the planar field. In the case of planar estimation, camera position and rotation are obtained directly from homography matrix~\citep{hartley2003multiple}. Therefore, to correctly project $\mathbf{l}$ onto the image, $\mathbf{p}$ and $\mathbf{q}$ must be checked and corrected to satisfy the next planar condition:
\begin{equation}
  \mathbf{q_c} =
  \begin{dcases}
     \mathbf{q} + \frac{\mathbf{n_t}\cdot(\mathbf{t}-\mathbf{q})}{\mathbf{n_t}\cdot(\mathbf{p}-\mathbf{q})} + \epsilon(\mathbf{n_t}) & \text{if $(\mathbf{q}-\mathbf{t})\cdot\mathbf{n_t} < 0$} \\
     \mathbf{q} &\text{otherwise}
  \end{dcases},
\end{equation}
with $\epsilon(\mathbf{n_t})$ being a threshold added in the direction of the normal vector to ensure that $\mathbf{q_c}$ lies ahead of the camera plane $\mathbf{\Pi_c}$. ``$\cdot$'' denotes a dot product. Once the projected extremities are corrected, the line $\mathbf{l}$, and consequently $\bar{\mathbf{p}}$ and $\bar{\mathbf{q}}$, can be computed. Given the image observations $\bar{\mathbf{p}}_d$ and $\bar{\mathbf{q}}_d$, we then define the point-to-line distances, following~\citep{spain2007analytical}, to the projected line $\mathbf{l}$ as follows:
\begin{equation}
    \begin{aligned}
        d(\mathbf{l},\mathbf{l}_d) &= d(\mathbf{l},\bar{\mathbf{p}}_d) + d(\mathbf{l},\bar{\mathbf{q}}_d)\\
        &=\frac{|\Delta_y\bar{\mathbf{p}}_{d,x} - \Delta_x\bar{\mathbf{p}}_{d,y} + \bar{\mathbf{q}}_x\bar{\mathbf{p}}_y - \bar{\mathbf{q}}_y\bar{\mathbf{p}}_x|}{\sqrt{\Delta_y^2 + \Delta_x^2}}\\
        &+\frac{|\Delta_y\bar{\mathbf{q}}_{d,x} - \Delta_x\bar{\mathbf{q}}_{d,y} + \bar{\mathbf{q}}_x\bar{\mathbf{p}}_y - \bar{\mathbf{q}}_y\bar{\mathbf{p}}_x|}{\sqrt{\Delta_y^2 + \Delta_x^2}}
    \end{aligned},
\end{equation}
where $\Delta_y = \bar{\mathbf{q}}_y - \bar{\mathbf{p}}_y$ and $\Delta_x = \bar{\mathbf{q}}_x - \bar{\mathbf{p}}_x$. It is important to note that by representing lines using their endpoints, we obtain comparable error representations for both points and lines. Given an image detection $\bar{\mathbf{x}}_d$ of a predefined keypoint, we define a reprojection error as:
\begin{equation}
    d(\bar{\mathbf{x}}_d, \mathbf{x}) = \bar{\mathbf{x}}_d - \bm{\pi}(\bm{\Theta}, \mathbf{x}) ,
\end{equation}
where $\bm{\pi}(\bm{\Theta}, \mathbf{x})$ represents the projection of the keypoint world coordinate $\mathbf{x} \in \mathbb{R}^3$ onto the image plane, given a calibration estimate $\bm{\Theta}$. We can thus construct a unified cost function that integrates each of the error terms as follows:
\begin{equation}
    C = \alpha\sum_{i\in\mathcal{L}
    } d(\mathbf{l}_i,\mathbf{l}_{d,i}) + (1-\alpha)\sum_{j\in\mathcal{K}p_c} d(\bar{\mathbf{x}}_{d,j}, \mathbf{x}_j),
\label{eq:costfunction}
\end{equation}
where $\mathcal{L}$ represents the set of detected lines and $\mathcal{K}p_c = \mathcal{K}p \cup \mathcal{K}p_e \cup \mathcal{K}p_1 \cup \mathcal{K}p_2 \cup \mathcal{K}p_3$ denotes the set-theoretic union of all available keypoint sets. A weighting parameter, $\alpha$, is empirically set to balance the influence of points and lines in the cost function. Following~\citep{pumarola2017pl}, a recursive approach over the detected reprojection line and point error will be applied to optimize the pose parameters $\bm{\Theta}$ as a nonlinear least-squares problem. To mitigate line misdetection issues, each detected line is compared to its projected counterpart using the initial estimate $\bm{\Theta}$. A projection error threshold is applied to both extremities of the detected line before it is subsequently added to the $\mathcal{L}$ set. 

\section{Experiments}

This section provides an overview of the datasets we use, the evaluation metrics employed to assess our approach, as well as implementation details. Subsequently, we present both qualitative and quantitative results, comparing our method with state-of-the-art approaches and performing an ablation study of the keypoint sets effect on the method performance.

\subsection{Datasets}
To evaluate our method, we compare our results with state-of-the-art methods on the SoccerNet-Calibration~\citep{cioppa2022scaling}, the WorldCup~\citep{homayounfar2017sports} and the TS-WorldCup~\citep{chu2022sports} soccer datasets. 

\textbf{SN-Calib Dataset:} The SoccerNetV3-Calibration (SN23) dataset~\citep{cioppa2022scaling} comprises 22,816 images extracted from SoccerNet~\citep{deliege2021soccernet} videos and encompasses a broadcast-based multi-view nature, offering a broader range of camera perspectives beyond the main broadcast camera. \cite{cioppa2022scaling} provide annotations for all segments of the soccer field, encompassing lines, conics and goal posts. For each visible segment on the court, at least two annotated positions are provided, optimally representing the segment in a polyline format. For the conics drawn on the pitch, the annotations consist of a list of points that roughly give the circle shape when connected. Additionally, \cite{theiner2023tvcalib} provided manual camera view annotations for the SoccerNetv3-Calibration 2022 (SN22) test set version, allowing the creation of data subsets taking into account the camera view distribution. 

\textbf{WC14 Dataset:} The WorldCup 2014 dataset (WC14)~\citep{homayounfar2017sports} stands as the reference benchmark for sports field registration and consists of 209 images from ten games for training and 186 images from other ten games for testing and the corresponding manually annotated homography matrices from the FIFA WorldCup 2014. Additionally, \cite{theiner2023tvcalib} provides segment annotations in SN-Calib~\citep{cioppa2022scaling} format.

\textbf{TS-WC Dataset:} The TS-WorldCup dataset (TSWC)~\citep{chu2022sports} contains detailed field markings on 3,812 field images from 43 videos of Soccer WorldCup 2014 and 2018 in a time-sequence fashion, which is ten times larger than the WorldCup 2014 dataset.

\subsection{Evaluation Metrics}
The quality of estimated camera parameters or homography matrices can be evaluated both in 2D and 3D spaces.

\textbf{Jaccard Index ($\text{JaC}_\gamma$):} Following \cite{magera2024universal} benchmarking protocol, the evaluation relies on calculating the reprojection error between each annotated point and the line to which it belongs. Adopting a binary classification approach, each pitch segment is treated as a single entity. To be considered correctly detected, all points within the segment must have a reprojection error smaller than a threshold. The projection of pitch elements from densely sampled points of the soccer field 3D model yields a polyline for each segment. Therefore, a polyline representing a soccer field segment $s$ is classified as a true positive (TP) if $\forall p \in s: \min \left(d(p, \hat{s})\right) < \gamma$, being $\hat{s}$ the corresponding annotated segment and $\gamma$ the distance threshold in pixels. Otherwise, this segment is counted as a false positive (FP). Segments only present in the annotations are counted as false negatives (FN). Hence, the Jaccard Index for camera calibration, $\text{JaC}_\gamma$, at a threshold of $\gamma$ pixels is defined as:
\begin{equation}
    \text{JaC}_\gamma = \frac{\text{TP}_\gamma}{\text{TP}_\gamma + \text{FN} + \text{FP}} ,
\end{equation}
where it serves as a measure of calibration accuracy. We also measure the completeness rate (CR) as the number of camera parameters provided divided by the number of images with more than four semantically labeled lines in the dataset. The final score (FS) as an evaluation criterion is calculated as the product of CR and $\text{JaC}_5$.

\textbf{Intersection over Union:} The intersection over union (IoU) includes two components. First, $\text{IoU}_{part}$ quantifies the visible area of the video frame by warping that frame using both the estimated homography and the GT one, projecting them onto the template, and then calculating the IoU. Second, $\text{IoU}_{whole}$ evaluates the entire sports field by warping the template with the refined homography, projecting it onto the original template, and calculating the IoU.

\textbf{Projection Error:} The projection error was quantified as the average distance, in meters, between the projected points using the estimated homography and the corresponding GT. To achieve that, we uniformly sampled 2,500 pixels from the visible field area of the camera image and projected them onto the field to compute the distance. The standard dimensions of a soccer field are 105$\times$68 meters.

\textbf{Reprojection Error:} The reprojection error was calculated by averaging the distance between the reprojected points in the video frame, utilizing both the estimated and the GT homography.

\subsection{Implementation Details}
Due to the absence of publicly available results on the multiple-view SN23~\citep{cioppa2022scaling} distribution, we trained two models from scratch: Multi-view (MV) and Single-view (SV). The latter is trained on a data subset composed almost entirely of non-replay frames, ensuring a high percentage of central camera shots. We train separate networks for the keypoints and line extremities detection tasks on the SN23-train dataset~\citep{cioppa2022scaling}. For the MV model, we train for $200$ epochs, using an initial learning rate of $1e^{-2}$ and a batch size of $2$. For the SV model, we train for $200$ epochs, using an initial learning rate of $1e^{-5}$ and a batch size of $1$. We utilize the Adam optimizer with default parameters $\beta_1 = 0.9$ and $\beta_2 = 0.999$. ${l}_2$-norm loss is used for heatmap regression in both neural networks. Data augmentation such as random horizontal flip, color jitter, and Gaussian noise are applied to enhance model robustness and generalization. Furthermore, we fine-tune both SV networks, keypoint and line extremity detection, on the WC14~\citep{homayounfar2017sports} and TSWC~\citep{chu2022sports} datasets. GT homographies are transformed into our proposed keypoint sets and line extremities by projecting their respective world coordinates to the field's ground plane. Note that non-planar points and lines are excluded from the transformation process, and their corresponding output layers are masked in the loss function. In fine-tuning, input images are resized to match the SoccerNet dataset image size. The experiments are conducted on a single NVIDIA GeForce RTX 2080 Ti GPU with $12$ GB of memory, and the implementation is carried out in the PyTorch framework. Calibration and further optimization tasks are conducted on a Intel Xeon Silver 4214 Processor and the implementations are carried out in the opencv-python and scipy frameworks, respectively.

\subsection{Results and comparisons}
We now present the results of an extensive evaluation, divided into camera calibration and homography estimation. The camera calibration evaluation assesses the accuracy of individual camera parameters using the $\text{JaC}_\gamma$ metric, while the homography estimation is evaluated using the IoU one, projection error, and reprojection error. As previously noted, methods with the subscript MV and SV correspond to multi-view and single-view models, respectively, and the PnL designation indicates the use of point and line calibration refinement module.

\subsubsection{Camera Calibration}
In team sports such as soccer, the action takes place on a nearly planar field. Consequently, most methods utilize homography estimation to map all elements positioned on this plane but cannot project non-planar points such as points belonging to goal posts or crossbars. Conversely, in~\citep{theiner2023tvcalib,mavrogiannis2024using,falaleev2024enhancing} make use of a 3D model of the soccer field to extract camera pose and intrinsic parameters directly. In homography-based approaches, parameter retrieval is accomplished through homography decomposition (HDecomp). We conduct a quantitative comparison of our proposed method to state-of-the-art approaches~\citep{chen2019sports, jiang2020optimizing, theiner2023tvcalib, mavrogiannis2024using} on the SN22-test-center dataset, comprising only images where the main camera center is visible (1,454 images). Furthermore, utilizing the SoccerNet annotation format for the WC14-test dataset provided by \cite{theiner2023tvcalib}, we conduct a comparison of our proposed method's performance in camera parameter estimation on the WorldCup 2014 dataset distribution.

\begin{figure}[t!]
  \centering
  \includegraphics[trim= 125 120 100 150, clip, width=0.49\textwidth]{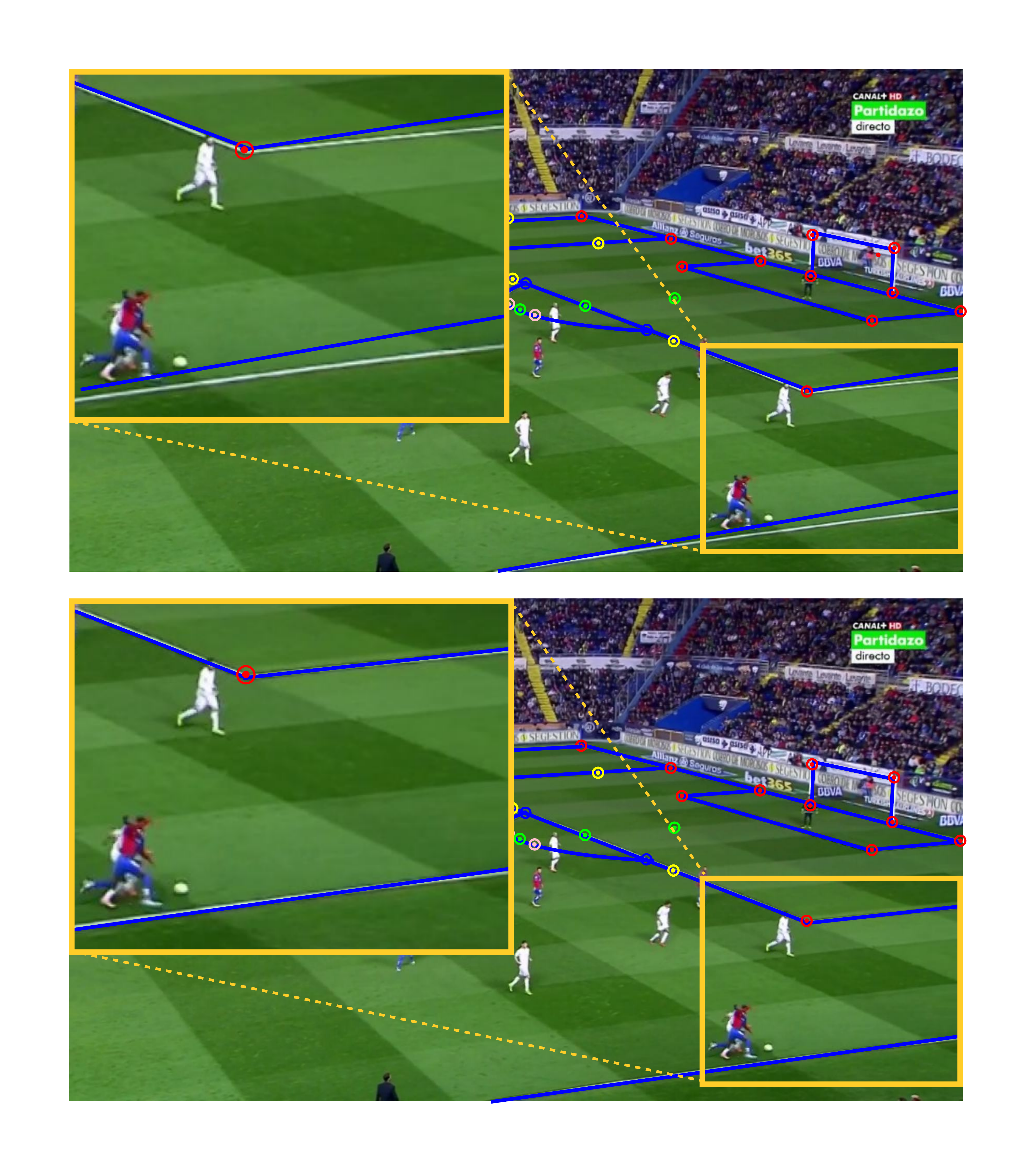}
  \caption{\textbf{Qualitative results of the PnL refinement module.} \textbf{Top:} Projection of soccer field lines with initial calibration estimate $\{\mathbf{K},\mathbf{R},\mathbf{t}\}$. \textbf{Bottom:} Projection of soccer field lines with refined calibration $\{\mathbf{K},\mathbf{R'},\mathbf{t'}\}$ through PnL refinement module.}
  \label{fig:qualitative2}
\end{figure}

We report the statistics from~\citep{theiner2023tvcalib} for the results of state-of-the-art approaches~\citep{chen2019sports, jiang2020optimizing, theiner2023tvcalib}, along with results provided by~\citep{mavrogiannis2024using}. As it is shown in Tables~\ref{tab:01}-\ref{tab:02} for the SN22-test and WC14 datasets, respectively, our SV method outperforms state-of-the-art approaches across several metrics in both datasets. Additionally, the inclusion of the PnL module demonstrates its effectiveness by increasing the FS metric by $5.7\%$ on the SN22-test-center dataset and a $8.3\%$ increase on the WC14-test one. Minor variations in CR, compared to previous approaches, stem from our method's requirement for a minimum number of visible keypoints for calibration and differences in the maximum allowable reprojection error to consider the parameter estimation valid. Moreover, variations in CR between our SV models are also due to the differing maximum allowable reprojection errors. Reprojection error is computed based on keypoint errors, meaning that while line-based refinement increases keypoint reprojection error, it results in improved overall calibration. Qualitative results demonstrating the effect of the PnL refinement module are presented in Fig.~\ref{fig:qualitative2}.

\begin{table}[t!]
\centering
\setlength{\tabcolsep}{3.5pt}
\renewcommand{\arraystretch}{1.1}
\normalsize
\begin{tabular}{l|l|ccc|c|c}
\specialrule{.15em}{0em}{.5em} 
                      &          & \multicolumn{3}{c}{\text{$\text{JaC}_\gamma$ {[}\%{]}}} &   &    \\
\text{Dataset}               & \text{Approach} & \text{5}      & \text{10}      & \text{20}     & \text{CR} & \text{FS} \\ \bottomrule \toprule
\multirow{3}{*}{\begin{tabular}[l]{@{}l@{}}SN22-test\\ -center\end{tabular}} & \citep{chen2019sports} + HDecomp  & 34.4 & 64.6 & 81.3 & 66.6 & 22.9\\
& TVCalib ($\tau$) \citep{theiner2023tvcalib} & 57.6 & 81.7 & 93.2 & 93.7 & 53.9 \\
& TVCalib \citep{theiner2023tvcalib} & 54.8 & 78.5 & 90.4 & \textbf{100.0} & 54.8\\
& \citep{mavrogiannis2024using} & 63.9 & 80.7 & 86.3 & \textbf{100.0} & 63.9\\ 
& Ours$_{\text{SV}}$ & 74.4 & 88.8 & 92.6 & 99.1 & 73.8\\
& Ours$_{\text{SV}}$ + PnL & \textbf{80.6} & \textbf{91.6} & \textbf{93.7} & 98.7 & \textbf{79.5}\\
\bottomrule \toprule
SN23-test               & \citep{falaleev2024enhancing} & 76.6 & - & - & 73.6 & 56.3 
                        \\
                        & Ours$_{\text{MV}}$        & 73.1 & 86.1 & 89.6 & \textbf{80.1} & 58.6 \\
                        & Ours$_{\text{MV}}$ + PnL     & \textbf{78.7} & \textbf{89.6} &  \textbf{91.9} & 78.4 & \textbf{61.8} \\
                        \bottomrule
\end{tabular}
\vspace{0.2cm}
\caption{\textbf{Evaluating camera calibration on SoccerNet~\citep{deliege2021soccernet} distributions.} The table reports our results for the Single-view model on the SN22-test-center dataset as well as for the Multi-view model on the full SN23-test dataset, including in both cases comparisons with competing approaches. For computing the $\text{JaC}_{\gamma}$ metric, we consider $\gamma=\{5, 10, 20\}$.}
\label{tab:01}
\end{table}

\begin{table}[t!]
\centering
\setlength{\tabcolsep}{7pt}
\renewcommand{\arraystretch}{1.1}
\normalsize
\begin{tabular}{l|ccc|c|c}
\specialrule{.15em}{0em}{.5em} 
   & \multicolumn{3}{c}{\text{$\text{JaC}_\gamma$ {[}\%{]}}} &    \\
\text{Approach}  & \text{5} & \text{10}    & \text{20}   & \text{CR} & \text{FS} \\ \bottomrule \toprule
\citep{chen2019sports} + HDecomp  & 32.7 & 67.3 & 87.3 & 81.7 & 26.7\\
\citep{jiang2020optimizing} + HDecomp & 36.9 & 66.4 & 83.9 & 84.9 & 31.3\\
TVCalib ($\tau$) \citep{theiner2023tvcalib} & 41.3 & 73.6 & 91.4 & 95.7 & 39.5\\
TVCalib \citep{theiner2023tvcalib} & 39.9 & 71.9 & 90.5 & \textbf{100.0} & 39.9\\
\citep{mavrogiannis2024using} & 59.4 & 82.3 & 90.9 & \textbf{100.0} & 59.4\\
Ours$_{\text{SV}}$ & 77.6 & 89.8 & 93.7 & \textbf{100.0} & 77.6 \\
Ours$_{\text{SV}}$ + PnL & \textbf{85.9} & \textbf{93.7} &  \textbf{95.8} & \textbf{100.0} & \textbf{85.9} \\\specialrule{.15em}{.5em}{0em} 
\end{tabular}
\vspace{0.2cm}
\caption{
\textbf{Quantitative comparison} of our Single-view model on camera calibration conducted on the WC14-test dataset. For computing the $\text{JaC}_{\gamma}$ metric, we consider $\gamma=\{5, 10, 20\}$.}
\label{tab:02}
\end{table}

\begin{figure}[t!]
  \centering
  \includegraphics[trim= 0 0 25 0, clip, width=0.49\textwidth]{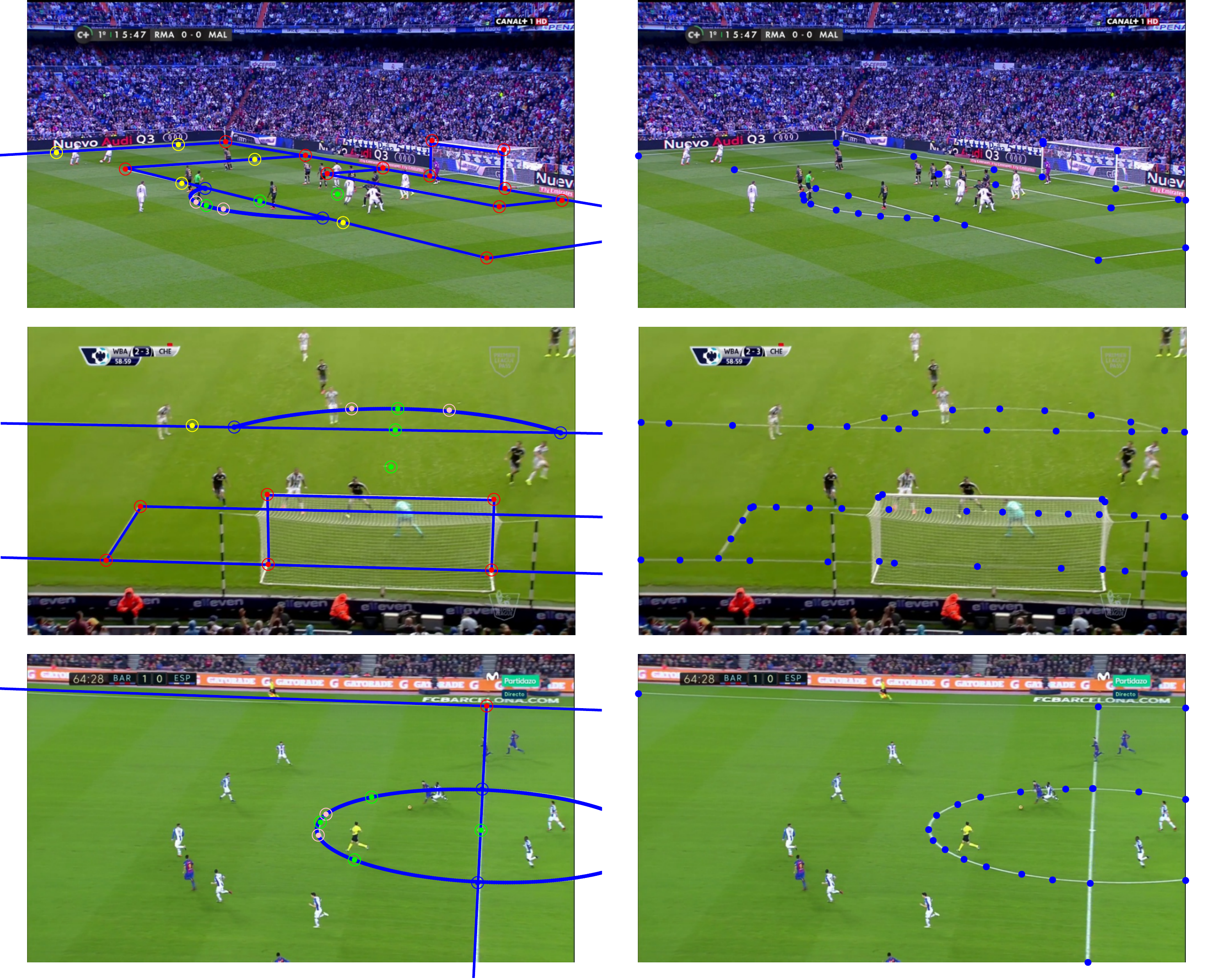}
  \caption{\textbf{Qualitative results of our MV model on SN23-test.} \textbf{Left:} Projection of soccer field lines and goal posts from world to image coordinates using predicted camera parameters. Blue lines correspond to segment projections, and colored points represent predicted keypoints (along with auxiliary points retrieved from line extremities detection). \textbf{Right:} SN23-test dataset annotations, where each soccer field line is delineated by a point set.}
  \label{fig:qualitative}
\end{figure}

\begin{figure}[h]
  \centering
  \includegraphics[trim= 125 120 100 130, clip, width=0.49\textwidth]{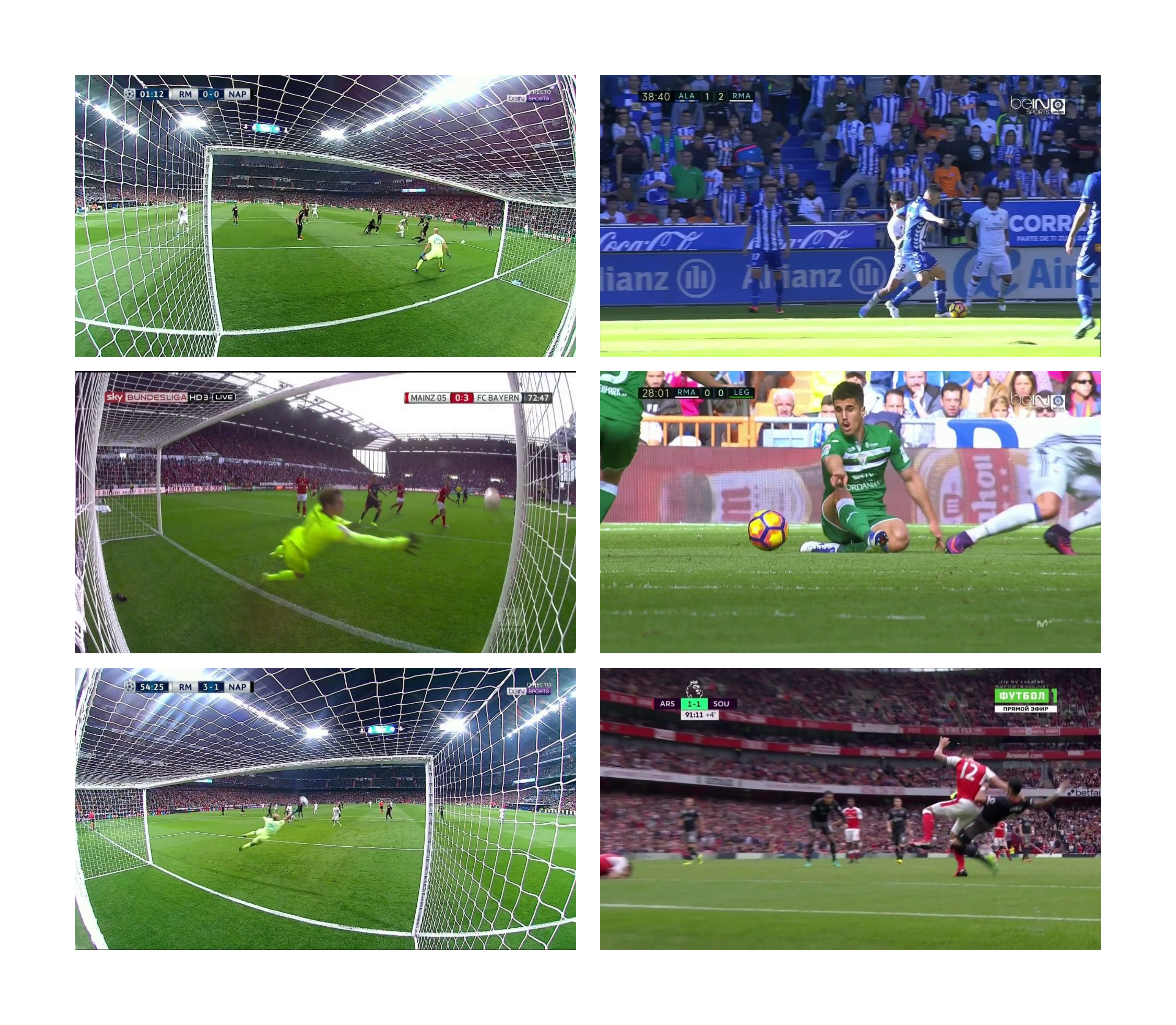} 
  \caption{\textbf{Challenging camera shots where calibration is unable.} \textbf{Left:} Fisheye shots from inside the goals with extreme lens distortion. \textbf{Right:} Close-up shots with few or none visible landmarks.}
  \label{fig:unable}
\end{figure}

We also evaluate our method on the entire SN23-test dataset, reporting statistics from~\citep{falaleev2024enhancing}, which represents the only comparable approach capable of extending calibration assessments to full multi-view scenarios. Although the calibration approach proposed by~\cite{mavrogiannis2024using} extends beyond the main camera, it is limited to calibrating the main, offside, and behind-the-goalpost cameras. Moreover, their method requires training a separate model for each camera location, whereas our approach generalizes keypoint and line extremity detection across the entire SoccerNet distribution using a single model. As shown in Table~\ref{tab:01} for the
SN23-test datasets, our MV model outperforms the existing approaches across all the presented metrics. Additionally, our PnL refinement module showcases a $3.2\%$ increase in the FS metric. The weighting parameter $\alpha$ in Eq.~\eqref{eq:costfunction} is empirically set to $\alpha = 0.6$ to maximize the FS metric across camera calibration experiments. An ablation study of the calibration results with respect to different values of $\alpha$ is presented in Table~\ref{tab:ablalpha}, evaluated on the SN23-test dataset. Qualitative results showcasing calibration results of our base pipeline on different camera angles are presented in Fig.~\ref{fig:qualitative}. The significant decrease in performance compared to the SN22-test-center dataset is attributed to the challenges of calibrating images with camera poses that deviate substantially from the main camera, such as close-up shots, where few or no landmarks are visible, or fisheye shots from inside the goals, where extreme lens distortion makes calibration unable. Examples of these challenging camera shots are shown in Fig.~\ref{fig:unable}.

In terms of inference speed, an ablation study on the frame rate is presented in Table~\ref{tab:abltime} under various configurations of heuristic voting, evaluated on the SN23-test set. Specifically, we assess different RANSAC configurations (ranging from no filtering to a $50$\,px threshold) and different keypoint subsets (i.e., \textit{full-keypoints}, \textit{main-keypoints}, and \textit{ground-plane-keypoints}). Using a fixed RANSAC threshold of $5$\,px and the \textit{full-keypoints} subset, our method initially achieves exhibits near-competitive performance at $164$\,ms, with the PnL optimization module accounting for approximately an $8\%$ increase in inference time. By evaluating six different RANSAC configurations with a single keypoint subset, the method achieves competitive calibration performance at $273$\,ms. State-of-the-art calibration accuracy is obtained when employing the full set of heuristic voting configurations, resulting in an inference time of 439\,ms. All reported timings are averaged over 10 independent runs, using the proposed image resolution and the software and hardware settings described above. It is also important to note that inference time can vary significantly across frames, primarily due to differences in the number of detected lines and the quality of initialization.

\subsubsection{Homography Estimation}
The proposed method is compared with respect to state-of-the-art approaches~\citep{chen2019sports, jiang2020optimizing, citraro2020real, zhang2021high, zhang2023four, nie2021robust, shi2022self, chu2022sports, oo2023residual, theiner2023tvcalib, maglo2023individual} using the WC14-test dataset. Additionally, our method is also compared to state-of-the-art approaches~\citep{chen2019sports, nie2021robust, chu2022sports, maglo2023individual, oo2023residual} using the TSWC-test dataset. For computing IoU-based metrics, projection error, and reprojection error, we adopt the approach outlined in~\citep{chu2022sports}. The dimensions of the soccer field template are set at $115 \times 74$ yards for a fair comparison with previous approaches.

In the evaluation on the WC14-test dataset, we report performance metrics based on the findings from the respective works, as shown in Table~\ref{tab:03}. For a fair comparison with~\citep{citraro2020real}, we use the results from their {\em ours-w/o-players} approach, as reported in their paper. Our fine-tuned model produces comparable results to previous approaches across all IoU-based, projection error, and reprojection error metrics. However, with the addition of the PnL refinement module, our method achieves state-of-the-art solutions across all metrics, further demonstrating the effectiveness of the proposed refinement module also in the homography estimation task.

For the evaluation on the TSWC-test dataset, we report the results for~\citep{nie2021robust, chu2022sports, maglo2023individual, oo2023residual} in Table~\ref{tab:03}, observing how our fine-tuned model with the PnL refinement module outperforms those methods across all metrics.

\begin{table*}[t!]
\centering
\small
\setlength{\tabcolsep}{5pt}
\renewcommand{\arraystretch}{1} 
\begin{tabular}{@{}l|l|cccccccc@{}}
\specialrule{.15em}{0em}{.5em} 
\multicolumn{1}{l|}{\multirow{2}{*}{\text{Dataset}}} & \multicolumn{1}{l|}{\multirow{2}{*}{\text{Approach}}}         & \multicolumn{2}{c}{$\text{IoU}_{\text{part}}$ $\uparrow$ (\%)}                           & \multicolumn{2}{c}{$\text{IoU}_{\text{whole}}$ $\uparrow$ (\%)}                          & \multicolumn{2}{c}{Proj. $\downarrow$ (m)}                   & \multicolumn{2}{c}{Reproj. $\downarrow$} \\ \cline{3-10}
\multicolumn{1}{l|}{}                         & \multicolumn{1}{c|}{} & \multicolumn{1}{c}{\size{8}{Mean}} & \multicolumn{1}{c}{\size{8}{Median}} & \multicolumn{1}{c}{\size{8}{Mean}} & \multicolumn{1}{c}{\size{8}{Median}} & \multicolumn{1}{c}{\size{8}{Mean}} & \multicolumn{1}{c}{\size{8}{Median}} & \multicolumn{1}{c}{\size{8}{Mean}} & \multicolumn{1}{c}{\size{8}{Median}} \\ \bottomrule \toprule
\multirow{2}{*}{WC14-test} 
                                & Jiang {\em et al.} \citep{jiang2020optimizing}  & 95.1 & 96.7 & 89.8 & 92.9 & - & - & - & -\\ 
                                & Citraro {\em et al.} \citep{citraro2020real} & - & - & 90.5 & 91.8 & - & - & 0.018 & 0.012\\
                                & Zhang {\em et al.} \citep{zhang2021high}& 95.9 & 97.5 & 91.4 & 94.2 & - & - & - & - \\
                                & Nie {\em et al.} \citep{nie2021robust} & 95.9 & 97.1 & 91.6 & 93.4 & 0.84 & 0.65 & 0.019 & 0.014 \\
                               & Shi {\em et al.} \citep{shi2022self} & 96.6 & 97.8 & 93.1 & 94.8 & - & - & - & - \\
                                & Chu {\em et al.} \citep{chu2022sports} & 96.0 & 97.0 & 91.2 & 93.1 & 0.81 & 0.63 & 0.019 & 0.014 \\
                                & Zhang et al \citep{zhang2023four}   & 95.9 & 97.3 & 91.4 & 94.1 & - & - & - & - \\
                               & Maglo {\em et al.} \citep{maglo2023individual}   & 96.3 & 97.4 & 92.0 & 94.1 & 0.74 & 0.55 & 0.018 & 0.014 \\
                               & Oo {\em et al.} \citep{oo2023residual}   & 96.9 & 97.9 & 92.9 & 94.6 & 0.65 & 0.46 & 0.016 & 0.012 \\ 
                               & Ours$_{\text{SV}}^{\ast\dagger}$ & 96.4 & 97.9 & 92.4 & 94.8 & 0.65 & 0.44 & 0.015 & 0.011\\
                               & Ours$_{\text{SV}}^{\ast\dagger}$ + PnL& \textbf{97.0} & \textbf{98.2} & \textbf{93.4} & \textbf{95.5} & \textbf{0.60} & \textbf{0.42} & \textbf{0.014} & \textbf{0.010}\\
                               \bottomrule \toprule
\multirow{2}{*}{TSWC-test}  
                               & Nie {\em et al.} \citep{nie2021robust}$^\ddagger$ & 97.4 & 97.8 & 92.5 & 94.2 & 0.43 & 0.38 & 0.011 & 0.010 \\
                              & Chu {\em et al.} \citep{chu2022sports}$^\ddagger$ & 98.1 & 98.2 & 94.8 & 95.4 & 0.36 & 0.33 & 0.009 & 0.008 \\
                              & Maglo {\em et al.} \citep{maglo2023individual}$^\ddagger$ & 98.3 & 98.5 & 95.7 & 96.2 & 0.26 & 0.23 & 0.008 & 0.006 \\
                              & Oo {\em et al.} \citep{oo2023residual}$^\ddagger$ & 98.5 & 98.7 & 95.8 & 96.7 & 0.26 & 0.21 & 0.007 & 0.006 \\ 
                              & Ours$_{\text{SV}}^{\ast\ddagger}$ & 98.2 & 98.4 & 94.6 & 95.8 & 0.28 & 0.24 & 0.007 & 0.006\\
                               & Ours$_{\text{SV}}^{\ast\ddagger}$ + PnL& \textbf{98.6} & \textbf{98.9} & \textbf{96.3} & \textbf{96.8} & \textbf{0.23} & \textbf{0.20} & \textbf{0.005} & \textbf{0.005}\\

\specialrule{.15em}{.5em}{.0em} 
\end{tabular}
\vspace{0.2cm}
\caption{\textbf{Evaluating the homography estimation on WC14-test and TSWC-test.} $\ast$ denotes the methods trained on SoccerNet distribution, $\dagger$ denotes the methods fine-tuned on the WC14 dataset and $\ddagger$ denotes the methods fine-tuned on the TSWC one.}
\label{tab:03}
\end{table*}

\subsubsection{Ablation Study on Keypoint Sets Contribution}
The contribution of each keypoint set, namely $\mathcal{K}p_{e}$, $\mathcal{K}p_{1}$, $\mathcal{K}p_{2}$, and $\mathcal{K}p_{3}$, is analyzed in Table~\ref{tab:abl}. Overall, the integration of each keypoint set results in improvements in the CR and FS metrics, with each set contributing to different geometric elements of the field. The inclusion of $\mathcal{K}p_{e}$ increases CR by providing visible landmarks along the main field's straight lines, particularly when line-line intersections are scarce in the image. However, despite these improvements in CR and FS, the Acc metrics decrease, as the method is able to calibrate more lines but with some lower-quality calibrations. A similar pattern is observed with the integration of the $\mathcal{K}p_{1}$ set, which boosts CR and FS by adding more visible landmarks to the field's straight lines and enabling the inclusion of field circles, though calibration quality in these areas remains suboptimal. A significant increase in CR, Acc, and consequently FS, is achieved through the integration of the $\mathcal{K}p_{2}$ set, particularly in frames where the field’s center circle is partially visible, but midfield line intersections are absent from the image. Finally, the integration of the $\mathcal{K}p_{3}$ set leads to further improvements across all metrics, enhancing the robustness of the method and setting a new state-of-the-art in sports field registration benchmarks.

\begin{table}[t]
\centering
\normalsize
\renewcommand{\arraystretch}{1}
\setlength{\tabcolsep}{5pt}
\begin{tabular}{cccc|ccc|c|c}
\specialrule{.15em}{0em}{.5em} 
   &    &    & \multicolumn{1}{l|}{}   & \multicolumn{3}{c}{\text{$\text{JaC}_\gamma$ {[}\%{]}}}     &       &    \\
$\mathcal{K}p_{e}$ & $\mathcal{K}p_{1}$ & $\mathcal{K}p_{2}$ & \multicolumn{1}{l|}{$\mathcal{K}p_{3}$} & \text{5} & \text{10} & \text{20} & \multicolumn{1}{l|}{\text{CR}} & \text{FS} \\
\bottomrule \toprule
\ding{55}& \ding{55} & \ding{55} & \ding{55} & 74.2 & 89.4 & 93.7 & 86.1 & 63.9\\
\ding{51}& \ding{55} & \ding{55} & \ding{55} & 74.0 & 89.0 & 93.5 & 88.8 & 65.7\\
\ding{51}& \ding{51} & \ding{55} & \ding{55} & 74.0 & 88.3 & 91.9 & 94.4 & 69.9\\
\ding{51}& \ding{51} & \ding{51} & \ding{55} & 75.1 & 89.2 & 92.6 & 97.8 & 73.5\\
\ding{51}& \ding{51} & \ding{51} & \ding{51} & 75.8 & 89.7 & 91.9 & 98.1 & \textbf{74.4} \\
\specialrule{.15em}{.5em}{.0em} 
\end{tabular}
\vspace{0.2cm}
\caption{\textbf{Ablation study of our keypoint sets.} The table shows the effect of every keypoint set on the SN22-test-center dataset.}
\label{tab:abl}
\end{table}

\begin{table}[t]
\centering
\normalsize
\renewcommand{\arraystretch}{1}
\setlength{\tabcolsep}{5pt}
\begin{tabular}{c|ccc|c|c}
\specialrule{.15em}{0em}{.5em} 
& \multicolumn{3}{c}{\text{$\text{JaC}_\gamma$ {[}\%{]}}}     &       &    \\
\multicolumn{1}{l|}{$\alpha$} & \text{5} & \text{10} & \text{20} & \multicolumn{1}{l|}{\text{CR}} & \text{FS} \\
\bottomrule \toprule
0.3 & 73.8 & 86.0 & 89.3 & \textbf{80.4} & 59.3 \\
0.4 & 75.5 & 86.9 & 89.9 & 79.7 & 60.2 \\
0.5 & 76.6 & 87.6 & 90.3 & 79.3 & 60.7 \\
0.6 & \textbf{78.8} & \textbf{89.6} & \textbf{91.9} & 78.4 & \textbf{61.8}\\
0.7 & 78.0 & 88.9 & 91.4 & 78.2 & 61.0 \\
\specialrule{.15em}{.5em}{.0em} 
\end{tabular}
\vspace{0.2cm}
\caption{Ablation study of the $\alpha$ parameter in the PnL optimization module, evaluated on the SN23-test dataset.}
\label{tab:ablalpha}
\end{table}

\begin{table*}[t]
\centering
\normalsize
\renewcommand{\arraystretch}{1}
\setlength{\tabcolsep}{5pt}
\begin{tabular}{ccc|c|c|c}
\specialrule{.15em}{0em}{.5em} 
$\#\text{RANSAC configs}$ & $\#\text{Keypoint subsets}$ & \text{PnL module} & $\text{Frame rate (ms)}\downarrow$ & CR & $\text{FS}\uparrow$ \\
\bottomrule \toprule
1 & 1 & \ding{55} & \textbf{164} & 74.6 & 52.7\\
1 & 1 & \ding{51} & 178 & 72.3 & 54.1 \\
6 & 1 & \ding{51} & 273 & 74.5 & 57.7\\
6 & 3 & \ding{51} & 439 & \textbf{78.4} & \textbf{61.8} \\

\specialrule{.15em}{.5em}{.0em} 
\end{tabular}
\vspace{0.2cm}
\caption{\textbf{Ablation study on the inference time of our method.} The table presents the method's frame rate under various heuristic voting configurations and with or without the use of the PnL module, evaluated on the SN23-test dataset.}
\label{tab:abltime}
\end{table*}

\subsubsection{Ablation Study on Refinement Contribution}
The contribution of each field geometrical object for initial estimate refinement—namely points, lines, and the full PnL module—is analyzed in Table~\ref{tab:abl2}. Similar behaviors are observed across the SoccerNet distributions: after obtaining the initial calibration estimate using the proposed keypoint sets, point-based refinement does not significantly alter the results. While point refinement slightly reduces reprojection error, leading to a minor increase in the CR metric, it simultaneously lowers accuracy metrics. In contrast, line-based refinement not only reduces reprojection error but also improves accuracy metrics, as reprojection error is calculated using points, and the maximum allowable reprojection error in the method affects these outcomes. Thus, line refinement generally produces more accurate calibration results. Finally, we confirm the previously noted effectiveness of the PnL module, which balances the contributions of both geometrical objects to deliver state-of-the-art results in the SoccerNet benchmarks.

For the WC14 dataset, a different behavior is observed: point-based refinement has no impact, and line-based refinement results in worse calibration across all metrics. Nevertheless, the proposed PnL refinement still achieves state-of-the-art results.

\begin{table}[h]
\centering
\normalsize
\renewcommand{\arraystretch}{1}
\setlength{\tabcolsep}{5pt}
\begin{tabular}{c|cc|ccc|c|c}
\specialrule{.15em}{0em}{.5em} 
\multicolumn{1}{l|}{\multirow{2}{*}{\text{Dataset}}}    &   &  \multicolumn{1}{l|}{}   & \multicolumn{3}{c}{\text{$\text{JaC}_\gamma$ {[}\%{]}}}  & & \\
 & P & L & \text{5} & \text{10} & \text{20} & \multicolumn{1}{l|}{\text{CR}} & \text{FS} \\
\bottomrule \toprule
\multirow{3}{*}{SN23-test} & \ding{55} & \ding{55} & 73.1 & 86.1 & 89.6 & 80.1 & 58.6 \\
& \ding{51} & \ding{55} & 71.6 & 84.4 & 88.1& 81.5 & 58.4 \\
& \ding{55} & \ding{51} & 75.8 & 88.0 & 92.3 & 72.4 & 54.9 \\
& \ding{51} & \ding{51} & 78.7 & 89.6 & 91.9 & 78.4 & \textbf{61.8}\\\hline
\multirow{3}{*}{\begin{tabular}[l]{@{}l@{}}SN22-test\\ -center\end{tabular}} & \ding{55} & \ding{55} & 74.4 & 88.8 & 92.6 & 99.1 & 73.8\\
& \ding{51} & \ding{55} & 74.4 & 88.9 & 92.7 & 99.1 & 73.8 \\
& \ding{55} & \ding{51} & 75.8 & 88.8 & 93.5 & 96.8 & 73.3\\
& \ding{51} & \ding{51} & 80.6 & 91.6 & 93.7 & 98.7 & \textbf{79.5}\\\hline
\multirow{3}{*}{WC14-test} & \ding{55} & \ding{55} & 77.6 & 89.8 & 93.7 & 100.0 & 77.6\\
& \ding{51} & \ding{55} & 77.8 & 89.6 & 93.3 & 100.0 & 77.8 \\
& \ding{55} & \ding{51} & 77.3 & 91.3 & 95.1 & 96.7 & 74.8 \\
& \ding{51} & \ding{51} & 85.9 & 93.7 & 95.8 & 100.0 & \textbf{85.9}\\
\specialrule{.15em}{.5em}{.0em} 
\end{tabular}
\vspace{0.2cm}
\caption{\textbf{Ablation study of points and lines contribution in the refinement module.} The table presents the impact of points and lines, both individually and jointly, on the calibration estimate refinement across the SN23-test, SN22-test-center and WC14-test datasets.}
\vspace{-0.2cm}
\label{tab:abl2}
\end{table}

\section{Conclusion}
In this paper, we introduce a novel open-source framework for 3D sports field registration. Our proposed pipeline adopts a minimalist approach by solely utilizing the geometric properties of the soccer field. We demonstrate superior performance in 3D camera calibration on SoccerNet and WorldCup 2014 datasets compared to state-of-the-art methods, while also achieving comparable results in homography estimation on WorldCup 2014 and TS-WorldCup datasets. Additionally, we introduce a novel calibration refinement module that leverages field landmarks, such as keypoints and lines, which, when integrated with our calibration pipeline, further enhance performance in 3D camera calibration benchmarks and achieve state-of-the-art results in homography estimation. We also extend our pipeline to multi-view camera calibration, establishing the state-of-the-art benchmark for multi-view broadcast-based camera calibration in soccer. Our method exhibits promising results, highlighting the effectiveness of utilizing a robust field template without the need for further refinements. Furthermore, the refinement module showcases its effectiveness showing superior performance and improving calibration results when visible field landmarks are scarce. As long as the video frame distortions are not too harsh (i.e., fisheye camera shots) and a minimum of four keypoints are visible, the proposed sports field registration approach is shown to be effective. In future work, we plan to enhance our approach by incorporating temporal consistency across consecutive video frames, better aligning with the nature of sports video broadcasts. Additionally, a potential extension of this work includes integrating distortion modeling into the camera calibration pipeline, enabling support for a broader range of lenses—from minor distortions to fisheye effects.\\
\appendix
\section{Analytical Validation of Tangent Points}
\label{appendix_analytical_validation}

Consider an estimated homography matrix $\mathbf{H} \in \mathbb{R}^{3 \times 3}$, a conic represented in image coordinates by the symmetric matrix $\mathbf{\bar{c}}$, and an external point $\mathbf{\bar{p}}$. We compute the tangent lines from the external point $\mathbf{\bar{p}}$ to the conic $\mathbf{\bar{c}}$ in image coordinates. The corresponding contact points with the conic are denoted by $\mathbf{\bar{x}}$. Then, the point $\mathbf{\bar{x}}$ satisfies the following conditions:
\begin{itemize}
    \item $\mathbf{\bar{x}}^\top \mathbf{\bar{c}} \mathbf{\bar{x}} = 0$, i.e., the point $\mathbf{\bar{x}}$ lies on the conic $\mathbf{\bar{c}}$ in image coordinates.
    \item $\mathbf{\bar{p}}^\top \mathbf{\bar{c}} \mathbf{\bar{x}} = 0$, i.e., the tangent line to $\mathbf{\bar{c}}$ at $\mathbf{\bar{x}}$ passes through $\mathbf{\bar{p}}$.
\end{itemize}

The conic can be reprojected into world coordinates using the inverse homography as:
\[
\mathbf{c} = \mathbf{H}^{-\top} \mathbf{\bar{c}}\mathbf{H}^{-1}.
\]
We also define the world-coordinate representations of the point $\mathbf{\bar{p}}$ and the tangent contact point $\mathbf{\bar{x}}$ as:
\[
\mathbf{p} = \mathbf{H}^{-1} \mathbf{\bar{p}}, \quad \mathbf{x} = \mathbf{H}^{-1} \mathbf{\bar{x}}.
\]

\subsection*{Proof that $\mathbf{x}$ lies on the world conic $\mathbf{c}$}

We can show that $\mathbf{x}$ lies on $\mathbf{c}$ as follows:
\[
\begin{aligned}
\mathbf{x}^\top \mathbf{c} \mathbf{x} 
&= (\mathbf{H}^{-1} \mathbf{\bar{x}})^\top \mathbf{c} (\mathbf{H}^{-1} \mathbf{\bar{x}}) \\
&= \mathbf{\bar{x}}^\top \mathbf{H}^{-\top} \mathbf{c} \mathbf{H}^{-1} \mathbf{\bar{x}} \\
&= \mathbf{\bar{x}}^\top \mathbf{\bar{c}} \mathbf{\bar{x}} = 0.
\end{aligned}
\]
Hence, if $\mathbf{x}^\top \mathbf{c} \mathbf{x} = 0$, the point $\mathbf{x}$ lies on the conic $\mathbf{c}$.

\subsection*{Proof that line $\mathbf{p}\mathbf{x}$ is tangent to $\mathbf{c}$ at $\mathbf{x}$}

We can also show that the line joining $\mathbf{p}$ and $\mathbf{x}$ is tangent to the conic $\mathbf{c}$ at $\mathbf{x}$:
\[
\begin{aligned}
\mathbf{p}^\top \mathbf{c} \mathbf{x} 
&= (\mathbf{H}^{-1} \mathbf{\bar{p}})^\top \mathbf{c} (\mathbf{H}^{-1} \mathbf{\bar{x}}) \\
&= \mathbf{\bar{p}}^\top \mathbf{H}^{-\top} \mathbf{c} \mathbf{H}^{-1} \mathbf{\bar{x}} \\
&= \mathbf{\bar{p}}^\top \mathbf{\bar{c}} \mathbf{\bar{x}} = 0.
\end{aligned}
\]

Therefore, if $\mathbf{p}^\top \mathbf{c} \mathbf{x} = 0$, the line $\mathbf{p}\mathbf{x}$ is tangent to the conic $\mathbf{c}$ at the point $\mathbf{x}$.

\vspace{0.5cm}
\noindent \textbf{Acknowledgment.} This work has been supported by the project GRAVATAR PID2023-151184OB-I00 funded by MCIU/AEI/10.13039/501100011033 and by ERDF, UE and by the Government of Catalonia under Joan Oró FI 2024 grant.

\bibliographystyle{splncs04}
\bibliography{main}
\end{document}